%% file: main.tex
\pdfoutput=1

\documentclass[11pt]{article}

\usepackage[]{ACL2023}

\usepackage{times}
\usepackage{latexsym}

\usepackage[T1]{fontenc}

\usepackage[utf8]{inputenc}

\usepackage{microtype}

\usepackage{inconsolata}

\title{Diverse Demonstrations Improve In-context Compositional Generalization}

\author{Itay Levy$^*$ ~~~~~~Ben Bogin$^*$ ~~~~~~
Jonathan Berant \\
The Blavatnik School of Computer Science, Tel-Aviv University \\
\texttt{\{itay.levy,ben.bogin,joberant\}@cs.tau.ac.il}}

\input{commands}
\usepackage{colortbl}         
\usepackage{amssymb}
\usepackage{amsmath}
\usepackage{amsfonts}
\usepackage{pifont}
\usepackage{tikz}
\usetikzlibrary{tikzmark}
\usepackage[linesnumbered,ruled,vlined,noend]{algorithm2e}
\usepackage{multirow}
\usepackage{threeparttable}
\usepackage{booktabs}
\usepackage{makecell}
\usepackage{xcolor}
\usepackage{graphicx}

\begin{document}
\maketitle

\def\thefootnote{*}\footnotetext{Equal contribution}\def\thefootnote{\arabic{footnote}}
\input{0_abstract}
\input{1_intro}
\input{2_background}
\input{3_method}
\input{4_experiments}
\input{5_related}
\input{6_conclusion}
\input{7_limitations}
\input{8_acknowledgments}

\bibliography{anthology,custom}
\bibliographystyle{acl_natbib}

\clearpage
\appendix
\input{app_a}

\end{document}

%% file: commands.tex
\newif\ifcomments
\commentstrue
\ifcomments
    \providecommand{\bb}[1]{{\protect\color{olive}{[BB: #1]}}}
    \providecommand{\jb}[1]{{\protect\color{red}{[JB: #1]}}}
    \providecommand{\il}[1]{{\protect\color{cyan}{[IL: #1]}}}
\else
    \providecommand{\bb}[1]{}
    \providecommand{\jb}[1]{}
    \providecommand{\il}[1]{}
\fi

\newcommand{\tightparagraph}[1]{\smallbreak\noindent\textbf{#1}}

%% file: 0_abstract.tex
\begin{abstract}
In-context learning has shown great success in i.i.d semantic parsing splits, where the training and test sets are drawn from the same distribution. In this setup, models are typically prompted with demonstrations that are \emph{similar} to the input utterance. However, in the setup of compositional generalization, where models are tested on outputs with structures that are absent from the training set, selecting similar demonstrations is insufficient, as often no example will be similar enough to the input. In this work, we propose a method to select \emph{diverse} demonstrations that aims to collectively \emph{cover} all of the structures required in the output program, in order to encourage the model to generalize to new structures from these demonstrations.
We empirically show that combining diverse demonstrations with in-context learning substantially improves performance across three compositional generalization semantic parsing datasets in the pure in-context learning setup and when combined with finetuning.\footnote{Our code is available at: \url{https://github.com/itayle/diverse-demonstrations}}
\end{abstract}

%% file: 1_intro.tex
\section{Introduction}
Despite strong performance of pretrained language models (LMs) across many tasks, they have been shown to struggle in a compositional generalization setting \cite{pmlr-v80-lake18a,Furrer2020CompositionalGI,shaw-etal-2021-compositional}, when tested on their ability to process and generate novel combinations of previously observed elements. For example, a model might fail to interpret the request \emph{``Book a meeting with Jake's supervisor''} even when \emph{``Book a meeting with Jake''} and \emph{``Who is Jake's supervisor?''} were observed during training. 
In semantic parsing, the task of mapping natural language utterances to formal queries, such generalization is important (especially in a real-world setting), since models are required to interpret new combinations that are not covered by the annotated training data \cite{herzig-berant-2019-dont,yin-etal-2021-compositional}.

\begin{figure}[!t]
  \centering
  \includegraphics[width=0.85\linewidth]{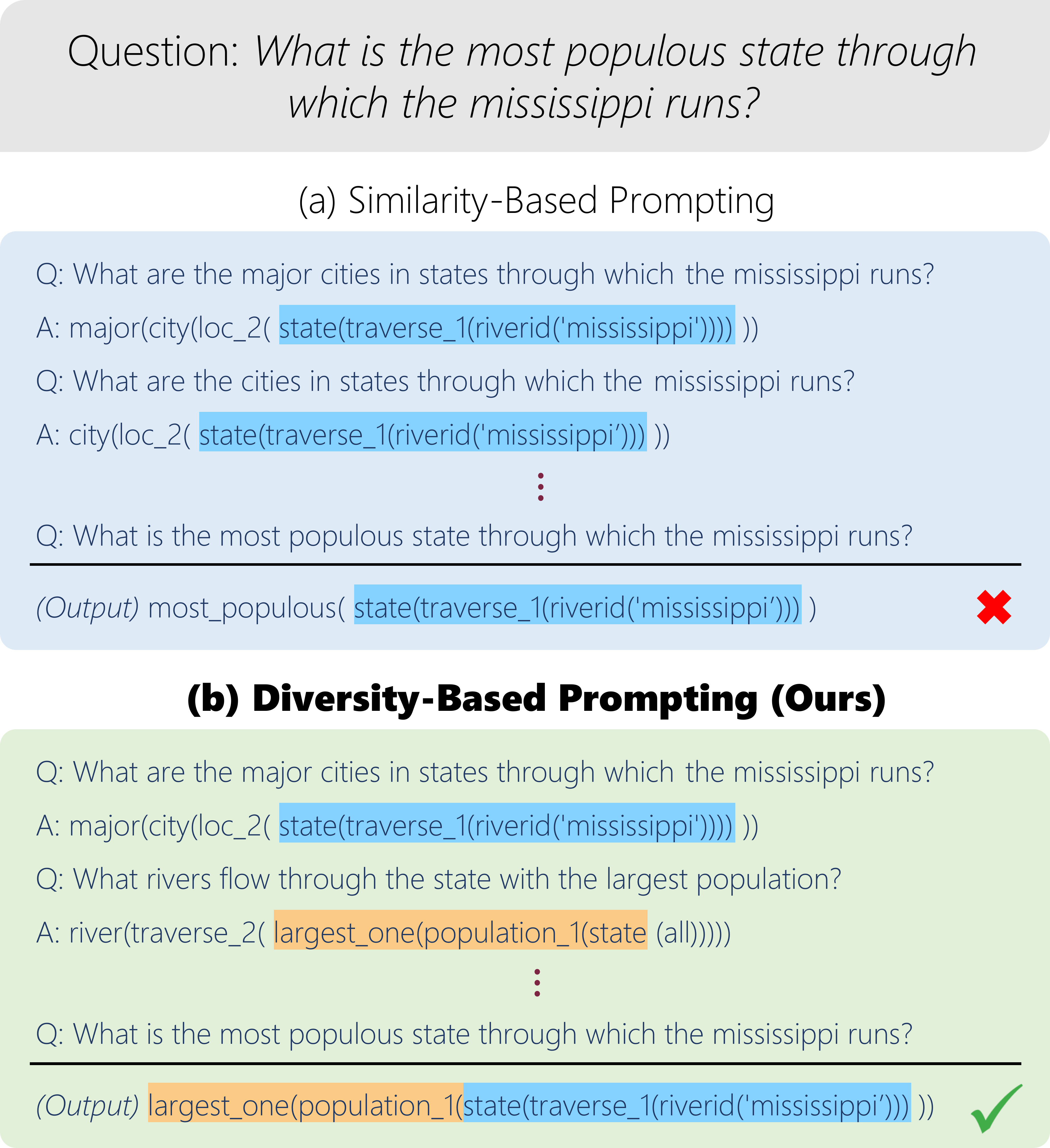}
  \setlength{\belowcaptionskip}{-5pt}
  \caption{Compositional generalization setup: (a) Selecting demonstrations by considering only similarity to the input yields repetitive demonstrations that do not cover the structures in the target program. (b) However, choosing diverse demonstrations enables better coverage and leads to a correct prediction.
  }
  \label{fig:intro}
\end{figure}
\begin{figure*}[!ht]
  \centering
  \includegraphics[width=0.9\textwidth]{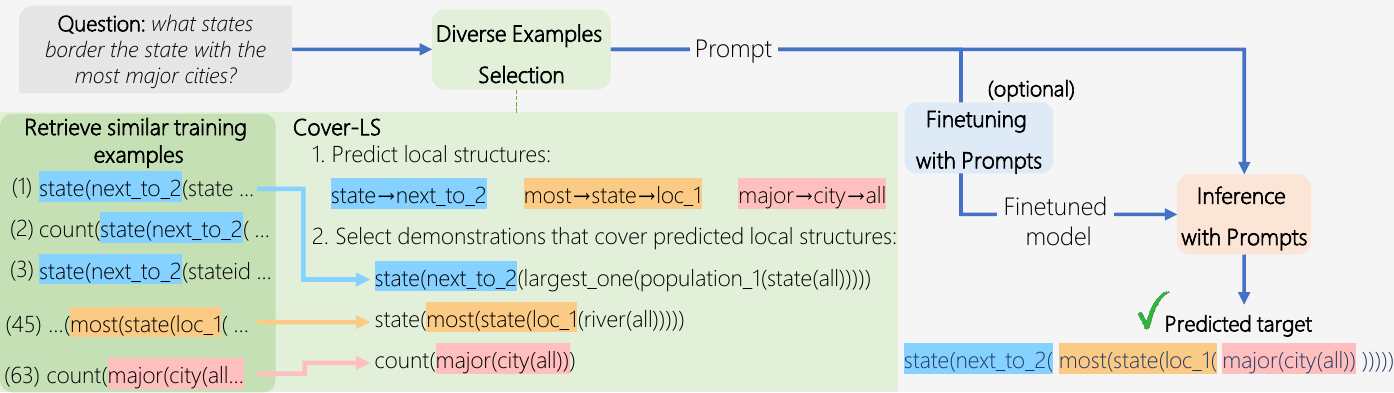}
  \caption{Overview of our framework. 
  Given an utterance, we construct a prompt by selecting a set of diverse demonstrations. Feeding the prompt to the model yields the predicted target. Optionally, models can be finetuned (FT setup). In the bottom left corner, we see how \emph{Cover-LS} selects diverse examples: predicting and covering \emph{local structures}, thereby enabling the selection of complementary examples.}
  \label{fig:overview}
\end{figure*}
Recently, large LMs have shown impressive performance on downstream tasks by conditioning on a text-based prompt that contains a few training examples.
This type of few-shot inference is known as \emph{in-context learning} (ICL, \citealp{Brown2020LanguageMA}). A core component of in-context learning is the set of examples in the prompt, often termed task \emph{demonstrations}.
With the right demonstrations, ICL can be an effective approach to improving LMs' compositional generalization abilities 
\cite{qiu-etal-2022-evaluating}.

Selecting a relevant set of demonstrations is crucial for generalization. However, most past work only considered the relevance of each example \emph{in isolation}, ignoring the quality of the entire set of examples \cite{liu-etal-2022-makes}. For instance, a retriever can be used to select the examples most similar to the input \cite{rubin-etal-2022-learning}. 
A set of demonstrations that are all highly relevant but highly similar to one another may not be as effective as a more \textbf{\emph{diverse}} set. In compositional splits, where no single demonstration is sufficiently similar to the input, choosing diverse demonstrations can be especially beneficial since it leads to better coverage of structures in the target program (Fig. \ref{fig:intro}). 

In this paper, we study how to leverage ICL to improve compositional generalization for semantic parsing, by optimizing the entire set of demonstrations and increasing the diversity of examples in this set.
We investigate two approaches for increasing diversity: (a) a \emph{coverage-based} approach, where we define a set of elements conditioned on the input utterance, and select examples that cover those elements (e.g., covering potential sub-structures in the output program), and (b) a second approach, where we select a subset of examples that are most dissimilar from one another, such that diversity is independent of the input utterance. Empirically, we find that coverage-based diversity results in better performance.

Our method can be used in the ``pure'' in-context learning setup without finetuning, which leverages the ability of large LMs, such as Codex \cite{Chen2021EvaluatingLL}, to generalize from the selected diverse demonstrations. Furthermore, it can be combined with finetuning by training a model with demonstrations as part of the input. This can be viewed as meta-learning, where the model learns to use demonstrations during training and build new structures based on them during inference
\cite{Finn2017ModelAgnosticMF,Lake2019CompositionalGT,conklin-etal-2021-meta,min-etal-2022-metaicl,chen-etal-2022-meta}. 
It can, however, lead to an over-reliance on demonstrations, especially in compositional splits. We address this by using ``noisy'' demonstrations during training.

We empirically test our method on three compositional generalization semantic parsing datasets. We show that diverse demonstrations, both with and without finetuning, improve performance by up to 23 absolute points (e.g., 50.3 $\rightarrow$ 73.5 on SMCalFlow-CS) compared to a baseline that retrieves demonstrations according to similarity alone, and lead to state-of-the-art results in multiple compositional setups.
Finally, we show that our method reduces the number of demonstrations needed for generalization and improves test performance on hard examples.

%% file: 2_background.tex
\section{Diversity for Compositional Generalization}

In semantic parsing, we define compositional splits of datasets as splits where train and test programs do not overlap \cite{finegan-dollak-etal-2018-improving}.
Recent work has shown that increasing the number of different program structures a model sees during training improves performance on compositional splits. This can be done by augmenting the training set 
\cite{qiu-etal-2022-improving} or through efficient sampling of diverse examples \cite{oren-etal-2021-finding,bogin-etal-2022-unobserved,gupta-etal-2022-structurally}. While past work focused on increasing structure diversity in the \emph{training set}, we focus on diversity in the \emph{demonstration set} within an ICL setup.

Increasing diversity is important as we want the demonstrations to \emph{cover} all structures of the expected output program. 
In the few-shot setting, where the model is unfamiliar with the formal language of the output programs, increasing coverage also improves generalization simply since otherwise the model will be unaware of the required program symbols (predicates and logical operators). 
However, selecting demonstrations that cover larger \emph{structures} (sub-trees of the program tree) are potentially more beneficial, for two reasons: (1) it reduces the amount of new structures that the model needs to produce, making demonstration fusion easier, and (2) it exposes the model to structure compositions in different contexts, providing the model with valuable information about how structures can be composed in the data.

%% file: 3_method.tex
\section{Diverse Demonstrations Selection}

\tightparagraph{Problem setup}
Given a training set $\mathcal{T}=\{(x_i,y_i)\}_{i=1}^n$ containing utterance-program pairs and a test utterance $x_{\small\text{test}}$, 
our objective is to select a subset of training examples $\mathcal{D}=\{(x_j,y_j)\}_{j=1}^k\subset\mathcal{T}$, where $k \ll n$, termed demonstrations.
Those demonstrations are then formatted as a text-based prompt $P$.
When feeding the concatenation of the prompt and the test utterance $([P;x_{\small\text{test}}])$  to the model, the desired output is $y_{\small\text{test}}$.

\tightparagraph{Overview}
Fig.~\ref{fig:overview} provides an overview of our framework for obtaining and leveraging diverse demonstrations for better compositional generalization.
Given an input utterance, $x_{\small\text{test}}$, we propose two approaches for selecting demonstrations.
In the first (\S\ref{sec:constrained_diversity}), we optimize \emph{coverage}: we define a set of elements that we want our demonstrations to cover (either structures in the program or utterance words), and then iteratively select examples that contain these elements. The second approach (\S\ref{sec:unconstrained_diversity}) increases diversity by selecting a subset of examples with minimal similarity. Fig.~\ref{fig:overview} shows an example of the former approach (\emph{Cover-LS}), where we predict and then attempt to cover \emph{local structures} (LS), i.e., sub-trees of the output program. Local structures were shown to be key for 
compositional generalization in \newcite{bogin-etal-2022-unobserved}.

Having selected demonstrations, we use them to construct a prompt (\S\ref{sec:prompt_construction}). We show that our method can be combined with finetuning to meta-train the model to learn in-context (\S\ref{sec:fine_tuning}).
 
\subsection{Coverage-based Selection}\label{sec:constrained_diversity}

\newcite{bogin-etal-2022-unobserved} have recently shown, in the context of finetuning semantic parsers, that models fail to generalize to programs with local structures that were not observed at training time, where local structures of a program are defined to be a set of its sub-trees.
Inspired by this observation, 
we propose \textbf{Cover-LS}, an algorithm that given the test utterance $x_{\small\text{test}}$, attempts to choose examples that collectively cover as many local structures as possible from the set $\mathcal{S}_{{y}_{\small\text{test}}}$ of local structures of the program $y_{\small\text{test}}$.
Since we have no access to $y_{\small\text{test}}$ at test time, we predict what local structures are likely using an auxiliary model, assuming that predicting local structures is \emph{easier} than predicting the entire program. Then, we iteratively select examples that cover the predicted local structures.

\tightparagraph{Local structures definition} 
We follow the definition of \newcite{bogin-etal-2022-unobserved}, and given a program $y$, convert it to its abstract syntax tree, where each tree node is a program symbol and parent-child edges connect functions to their arguments. In addition, we add ``sibling'' edges between consecutive arguments.
The local structures, $\mathcal{S}_{{y}_{\small\text{test}}}$, are a subset of all of the connected sub-graphs in the abstract syntax tree (e.g., \texttt{state$\rightarrow$next\_to\_2} and \texttt{most$\rightarrow$state$\rightarrow$loc\_1} in Fig.~\ref{fig:overview}, see more examples in Tab.~\ref{tab:ls_examples}), as defined in App.~\ref{app:local_structures}. Unlike \newcite{bogin-etal-2022-unobserved}, we consider local structures with any number of nodes.
In addition, we anonymize programs by replacing values such as strings and numbers with constants (\texttt{string} and \texttt{number}), since such values are usually not relevant for program coverage.

\tightparagraph{Predicting local structures}
As mentioned, we assume predicting local structures is easier than predicting an entire program. Thus, 
we train an auxiliary model by finetuning 
T5 \citep{Raffel2020ExploringTL} on the training set in the standard manner, training it to output anonymized programs given input utterances with no demonstrations.
Then, for each test utterance, $x_{\small\text{test}}$, we use beam search to output $B$ candidate programs $\{\tilde{y}_b\}_{b=1}^B$ and define the set of local structures as $\mathcal{S}_{\tilde{y}_{\small\text{test}}}=\bigcup_{b=1}^B \mathcal{S}_{\tilde{y}_b}$. 

\tightparagraph{Covering local structures}
Our goal is to choose a set of demonstrations, $\mathcal{D}$, that covers the local structures in $\mathcal{S}_{\tilde{y}_{\small\text{test}}}$.
Choosing an example for each local structure is infeasible due to prompt length limitations, and thus we propose 
Alg.~\ref{alg:cover_ls}, whose goal is to choose a small set of demonstrations that are (a) similar to the test utterance $x_{\small\text{test}}$ and (b) cover as many local structures in $\mathcal{S}_{\tilde{y}_{\small\text{test}}}$ as possible.

We sort the LSs based on their size (number of nodes) in descending order (line \ref{alg-line:sort}). By first selecting training examples with programs that contain \emph{larger} LSs from $\mathcal{S}_{\tilde{y}_{\small\text{test}}}$, we are more likely to include training examples similar to the test utterance, which should improve few-shot performance. Then, we iterate over all LSs, and for each local structure $s$ we \emph{retrieve} the most similar training example that contains $s$ (line \ref{alg-line:retrieve}), and add it to $\mathcal{D}$ (line \ref{alg-line:add}). We then update the pool of LSs such that it will include only LSs that are not yet covered (line \ref{alg-line:remove-ls}). To further encourage diversity, we remove from our example pool all examples that share the same template (program after anonymization) as the chosen examples (line \ref{alg-line:remove-anon}). We keep choosing examples until reaching the desired amount of demonstrations, which might result in choosing more than one example for each local structure (lines \ref{alg-line:while}-\ref{alg-line:reinit}). 

\begin{algorithm}[!t]
\small
\SetAlgoLined
\caption{Cover-LS Algorithm}\label{alg:cover_ls}
\SetKwInOut{Input}{Input}\SetKwInOut{Output}{Output}
\Input{List of candidate local structures to cover $\mathcal{S}$ ; 
Pool of training examples $\mathcal{T}$ ; 
Retriever $R$ ; 
Desired number of output examples $k$}
\Output{Set of training examples $\mathcal{D}$}
$\mathcal{D} = \emptyset$ \\
Sort $\mathcal{S}$ from largest to smallest \label{alg-line:sort}\\
\While{$|\mathcal{D}| < k$} 
{ \label{alg-line:while}
$\mathcal{S}_{\small\text{uncovered}}=\mathcal{S}$ \label{alg-line:reinit}\\
\For{each $s\in \mathcal{S}_{\small\text{uncovered}}$}  
{
	Retrieve with $R$ an example $e \in \mathcal{T}$ that contains $s$ \label{alg-line:retrieve}\\
	Add $e$ to $\mathcal{D}$ \label{alg-line:add}\\
	Remove from $\mathcal{S}_{\small\text{uncovered}}$ LSs that appear in $e$ \label{alg-line:remove-ls}\\
	Remove from $\mathcal{T}$ all examples with same anonymized program as $e$ \label{alg-line:remove-anon}\\
	\If{$|\mathcal{D}| == k$}{
   break
   }
}
}
\end{algorithm}

We assume (line \ref{alg-line:retrieve}) access to a retriever that takes as input an utterance and returns similar training examples, from which we filter only examples that contain the desired structure.
A variety of retrievers can be used, such as BM25 \cite{Robertson2009ThePR} or SBERT \citep{reimers-gurevych-2019-sentence}. 

We observe that in our setup, the running time of Cover-LS is negligible compared to the decoding time of the LMs.

\tightparagraph{Utterance coverage}
We propose a simpler variant that does not require predicting a set of local structures with an auxiliary model. This variant, termed \textbf{Cover-Utt}, uses the same coverage-oriented algorithm, but covers 
\emph{words} in the input utterance, rather than predicted local structures. This is beneficial when the quality of the auxiliary model, and consequently predicted LSs, is low.

\subsection{Diversity without Coverage}\label{sec:unconstrained_diversity}

The primary challenge with coverage-based approaches is identifying the elements that need to be covered.
An alternative approach is to define diversity more explicitly and select a subset of demonstrations that are dissimilar from one another (while being relevant for the input utterance).

A natural approach for choosing a subset of high-quality and diverse demonstrations from the training set is Determinantal Point Process (DPP)
\citep{Kulesza2012DeterminantalPP}, a probabilistic model that defines a probability distribution over subsets of items, giving high probability to subsets that contain \emph{relevant} and \emph{diverse} items.
DPP requires a \emph{relevance score} for each item and a \emph{similarity score} between pairs of items.
In our case, we define the relevance of a demonstration through its \emph{retriever score} for the input test utterance. To compute the similarity between demonstration pairs, we first extract LSs and compute tf-idf vectors for each demonstration. The similarity of each pair is then the cosine similarity between their tf-idf vectors. Full implementation details are in App.~\ref{app:dpp}.

\subsection{Prompt Construction}\label{sec:prompt_construction}
We order the chosen demonstrations according to their retriever score with respect to the input utterance in ascending order, in accordance to common practices \cite{liu-etal-2022-makes}. When finetuning the model (\S\ref{sec:fine_tuning}), demonstrations are shuffled.
Demonstrations are formatted to a prompt according to the format in App. \ref{app:prompt_format}, concatenated with the test utterance, and fed to the model.

\subsection{Finetuning with Prompts}\label{sec:fine_tuning}

Despite the success of ``pure'' in-context learning, where model parameters are frozen, it has been by and large restricted to very large LMs. Conversely, finetuning requires more training data, but performs well even with smaller models. In-context learning can be easily integrated with finetuning by training a model with demonstrations as part of the input. This paradigm can be considered as meta-learning, where the model learns how to use demonstrations during training \cite{min-etal-2022-metaicl}.

\input{tables/table_datasets}

When meta-learning is used in the i.i.d. setup, where the training and test examples are drawn from the same distribution, one can use the same procedure to select demonstrations at both training time and test time.
However, in a compositional generalization setup, this does not work:
at training time, the model will observe demonstrations that are similar to the target output and will learn to heavily rely on demonstrations and copy large chunks of them. Thus, the model will not learn to compose demonstration parts and will struggle with examples drawn from a different distribution.

To address this phenomenon, which we term \emph{over-copying}, past work \cite{pasupat-etal-2021-controllable,zemlyanskiy-etal-2022-generate} used \emph{sampling} to add noise to the demonstrations. Here, we also reduce the similarity of demonstrations to the input utterance, but with a simpler approach.
Recall that our Cover-LS algorithm picks similar examples by (a) finding demonstrations that share \emph{large} LSs with the predicted program (lines \ref{alg-line:sort}-\ref{alg-line:retrieve} in Alg.~\ref{alg:cover_ls}), and (b) using a retriever to find the most similar examples among these. To address over-copying, we modify this:
at training time, we only consider LSs of size 1, i.e., program symbols, and for each such LS we randomly choose an example that contains this symbol rather than use a powerful retriever. 

%% file: tables/table_datasets.tex
\begin{table*}[!t]

\centering
    \footnotesize
    \begin{threeparttable}
\begin{tabular}{@{}ll@{}}

\toprule
{\bf Dataset} & {\bf Example} \\
 \midrule
  \makecell[l]{\textbf{SMCalFlow-CS}
  } & \makecell[l]{\textit{Can you make a meeting with David Lax 's reports ?}\\ 
  \texttt{(Yield :output (CreateCommitEventWrapper :event (CreatePreflightEventWrapper} \\
  \texttt{:constraint (Constraint[Event] :attendees (AttendeeListHasPeople}  
  \\\texttt{:people (FindReports :recipient (Execute :intension (refer (extensionConstraint} \\ \texttt{(RecipientWithNameLike :constraint (Constraint[Recipient]) :name } \\ \texttt{\# (PersonName ``David Lax")))))))))))}}\\ 


\cmidrule[0.4pt](r{0.125em}){1-2}%
  \makecell[l]{\textbf{SMCalFlow-CS}\\\textbf{Simple}\\(natural)} 
  & \makecell[l]{\texttt{CreateEvent (with\_attendee  (FindReports (recipient= refer (Recipient?} \\\texttt{(name= LIKE (David Lax))))))}}\\ 
  
    \midrule
  \makecell[l]{\textbf{GeoQuery}\\(natural)} & \makecell[l]{\textit{What is the most populous state through which the mississippi runs ?} \\ \texttt{largest\_one (population\_1 (state (traverse\_1 (riverid ("mississippi")))))}}\\
  \midrule
 \makecell[l]{\textbf{COVR-10}\\(synthetic)} & \makecell[l]{\textit{What is the color of square dog ?} \\ \texttt{query\_attr[color] (filter (square, find (dog)))}}\\
 \bottomrule
\end{tabular}
\end{threeparttable}%
\caption{An example utterance-program pair for each of the datasets.
}
\label{tab:datasets}
\end{table*}

%% file: 4_experiments.tex
\section{Experiments}
\label{sec:experiments}

We present our experimental setup and results on different compositional semantic parsing tasks, with finetuning (FT) and without (NoFT). 
\input{tables/table_NFT}

\subsection{Datasets}
We evaluate our methods on three datasets (examples in Tab.~\ref{tab:datasets}).

\tightparagraph{SMCalFlow-CS}
is a few-shot compositional generalization dataset proposed by \citet{yin-etal-2021-compositional} derived from SMCalFlow \citep{andreas-etal-2020-task}. It contains single-turn natural sentences involving two domains (organization structure and event creation), each having its own set of program symbols. The test set of the compositional splits contains only cross-domain examples, where both domains appear. We show results for a few-shot setting (split $k$-C, where $k\in\{8,16,32\}$) where the training set includes only $k$ cross-domain examples, and a zero-shot setting (split 0-C). We also evaluate on an i.i.d. split\footnote{The split we use for the i.i.d. setup is 8-S.} where the test set contains only single-domain examples. Prior studies on the dataset employed LISP and LISPRESS program formats, resulting in v1 and v2 versions, respectively (see an example in Tab.~\ref{tab:app_smcal}). We default to using v1, unless otherwise specified.

For our FT experiments, we use \textbf{SMCalFlow-CS Simple}, which contains the same utterances as SMCalFlow-CS, but 
with programs that use a simplified syntax provided by \newcite{meron-2022-simplifying}. We opt for this version because programs are much shorter, leading to a smaller memory footprint and accelerating training and inference.

\tightparagraph{GeoQuery}
\cite{Zelle1996LearningTP,Tang2001UsingMC} contains 880 natural language questions about US geography. We use the standard (i.i.d.) and compositional splits created by
\citet{shaw-etal-2021-compositional}: (1) template split, where target programs are anonymized into templates and then the templates are randomly split between training and test sets \citep{finegan-dollak-etal-2018-improving}; (2) TMCD split, which makes the distributions of compounds in training and test sets as divergent as possible \cite{keysers2020measuring}; and (3) length split, where test sequences are longer than training ones.
Similar to prior work, we average results across three TMCD and template splits to reduce variance caused by the small dataset size.

\tightparagraph{COVR-10} COVR \cite{bogin-etal-2022-unobserved} is a synthetic dataset based on a variable-free functional language. COVR-10 contains 10 compositional grammar splits, in which each test set includes programs featuring a particular set of local structures not observed at training time. Results are averaged across the 10 splits.

\subsection{Experimental setup}
\tightparagraph{Models} We use Codex (code-davinci-002) \cite{Chen2021EvaluatingLL,Ouyang2022TrainingLM} for all NoFT experiments, and T5-large \citep{Raffel2020ExploringTL} for FT experiments. T5-large is used to predict LSs in both the NoFT and FT setups.

\tightparagraph{Evaluation} Like prior work, we use exact match accuracy as the main metric for evaluation. 
Results are averaged over 3 random seeds unless stated otherwise. In the FT setup, we use the entire test set for evaluation. In the NoFT setup, we use 100 test examples due to rate limits of the Codex inference API (and another 100 development examples for hyperparameter tuning).

\tightparagraph{Prompt} We use a prompt size of $k=24$ for NoFT experiments and $k=3$ for FT experiments, unless stated otherwise. A prompt is truncated when its length exceeds the model's context length (excluding the tokens reserved for generation). In FT experiments, we included only the programs in our demonstrations and discarded their utterances, due to limitations of memory and sequence length (preliminary experiments with utterances showed this does not affect accuracy).

\tightparagraph{Retrievers} 
In NoFT setup, we use BM25 over lower-cased utterance words. In FT setup, we use BM25 over predicted program symbols in $\mathcal{S}_{\tilde{y}_{\small\text{test}}}$ (predicted using T5).
In Cover-LS experiments we use a random retriever at training time to avoid over-copying.
We analyze other possible retriever choices in \S\ref{sec:retrivers}.

\tightparagraph{Hyperparameter tuning and model selection} We train two types of models in this work: (a) models for predicting LSs, and (b) models finetuned with prompts. 
For both cases, we use the development set whenever it is available for model selection, otherwise, we use the last checkpoint. Similarly, we use the development set to tune the number of beam candidates $B$ when predicting local structures, and if there is no development set, we set $B=1$. We detail finetuning hyperparameters in App. \ref{app:FT}.

\tightparagraph{Local structure size}
In some experiments, we limit the maximum size of local structures (the number of nodes they contain). A subscript notation ($\textrm{Cover-LS}_{d}$ or $\textrm{DPP}_{d}$) indicates a limit up to size $d$.

\subsection{Baselines}\label{sec:baselines}

\tightparagraph{Finetuning without prompts} Vanilla-finetuned T5 model which is trained without demonstrations, similar to the one used to predict LSs (\S\ref{sec:constrained_diversity}), except that it is trained on non-anonymized programs.

\tightparagraph{Top-K}
We construct the prompt with the top-$k$ examples that are most similar to $x_{\small\textrm{test}}$ according to the retriever score.

\tightparagraph{Random}
We construct a prompt by randomly sampling $k$ training examples without repetition.

We also conduct oracle experiments, where at test time we have access to $y_{\small\text{test}}$ both for retrieval and LS coverage. The retriever takes as input the gold program and scores demonstrations using BM25 over the gold program symbols. In oracle Cover-LS, we cover local structures from $\mathcal{S}_{{y}_{\small\text{test}}}$ without predicting them with a model.

\subsection{Main Results}
\input{tables/table_entire_test_alternative}
\tightparagraph{NoFT}
We observe (Tab.~\ref{tab:NFT}) that all methods for increasing diversity (Cover-Utt, DPP and Cover-LS) outperform Top-K, which selects similar demonstrations without accounting for diversity, in 7 out of 8 compositional splits.
In fact, all non-oracle diversity methods outperform an \emph{oracle} Top-K in 7 out of 8 compositional splits, suggesting that retrieval methods that only consider similarity are sub-optimal even in an oracle setup.
Similarly, all diversity methods improve performance compared to a finetuned T5 model in all compositional splits except GeoQuery's template splits.
Furthermore, sampling random examples (Random baseline) results in poor performance in GeoQuery and SMCalFlow-CS, but achieves high accuracy in COVR-10, beating all methods except Cover-Utt. This can be explained by the synthetic nature and small vocabulary of COVR-10.

Comparing diversity methods, Cover-LS and Cover-Utt are better than DPP in 7 out of 10 splits, showing that covering the target input/program goes beyond simply picking diverse examples. Cover-Utt, which covers utterance words, works surprisingly well considering its simplicity. Coverage-based methods also outperform Top-K in i.i.d splits. One noticeable failure of Cover-LS is the 0-C split, where it fails to generalize, due to the poor T5 performance on this split (T5 baseline gets 0 accuracy). This emphasizes that if one cannot reasonably predict LSs, then covering input words is a viable alternative.
Lastly, oracle methods 
outperform their non-oracle counterparts in most settings, but not always. This occurs because our oracle method, which has access to the gold program, does not guarantee the selection of the optimal set of demonstrations, a phenomenon also observed in \citet{qiu-etal-2022-evaluating}.
\begin{figure}[t!]
  \centering
  \includegraphics[width=\linewidth]{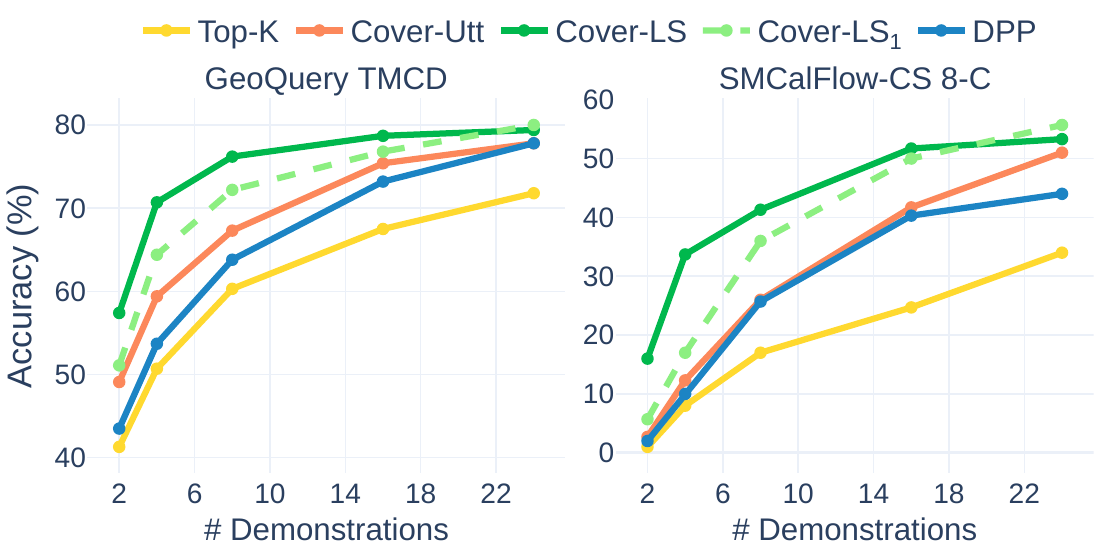}
  \caption{Comparing model accuracy (NoFT setup) based on the number of demonstrations, with multiple methods for selecting demonstrations.
  }
  \label{fig:prompt_size}
\end{figure}

Tab.~\ref{tab:entire_test_set} shows accuracy on the entire test set (NoFT setup). 
Since the underlying models differ substantially, a fair comparison to previous work is impossible. Nevertheless, a comparison still provides a high-level overview for the state of these tasks. Results show that using Codex with Cover-LS outperforms a T5 finetuned with augmentation \cite{qiu-etal-2022-improving} in 4 compositional splits out of 6 (TMCD, Length, 8-C and 32-C), and outperforms non-finetuned PaLM 540B, where demonstrations are selected using BM25, in all splits.

\paragraph{Number of demonstrations (NoFT)}
We examine how performance is affected by the number of demonstrations in Fig. \ref{fig:prompt_size}. Cover-LS outperforms Top-K by a large margin across all prompt sizes. Moreover, Cover-LS requires just four demonstrations in order to obtain roughly the same results as Top-K with 24 demonstrations. The gap between Cover-LS and Cover-Utt or $\textrm{Cover-LS}_{1}$ shows the importance of covering structures rather than just program symbols or utterance words, especially for small demonstration sets.

\input{tables/table_ft}

\tightparagraph{FT} Finetuning results are shown in Tab. \ref{tab:FT}, where we detail separately the method used for demonstration selection at both training time and test time, as those may diverge to avoid over-copying.

First, using random demonstrations at test time, without controlling for diversity or using any retriever, is better compared to using no demonstrations at all.
Our main method constructs prompts with Cover-LS at test time, but during training, prompts are retrieved with $\textrm{Cover-LS}_{1}$, that only covers program symbols, but not local structures, to avoid over-copying (see \S\ref{sec:fine_tuning}). This
combination leads to higher performance in all compositional splits compared 
to baselines that use Top-K or random sampling. Interestingly, using Top-K at both training time and test time yields low accuracy in compositional splits, but high results in i.i.d. splits. This corroborates our assumption that diversity is needed in compositional setups. Finally, A variant of our method, where $\textrm{Cover-LS}_{1}$ is used both during training and test time, is comparable to our main method across all splits.

We observe that limiting coverage at training time to program symbols is crucial: accuracy drops in all splits if we limit Cover-LS to structures up to size 2 ($\textrm{Cover-LS}_{2}$) instead of 1, or if we have no such limitation at all. The oracle Cover-LS outperforms all non-oracle models (unlike in NoFT, where this is not always the case).

\subsection{Analysis}\label{subsec:analysis}
\begin{figure}[!t]
  \centering
  \includegraphics[width=0.85\linewidth]{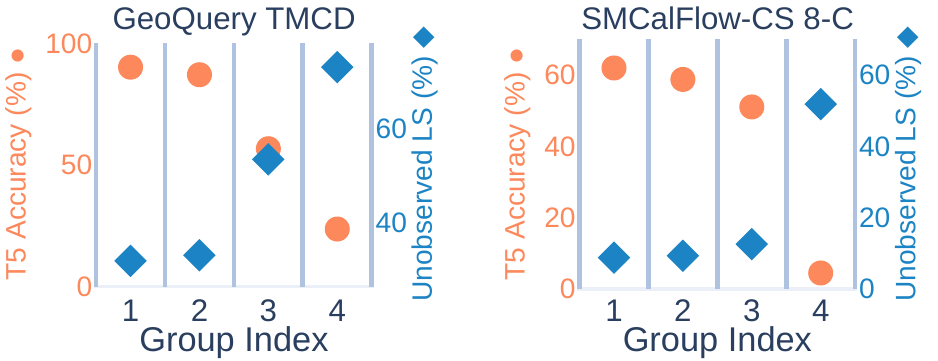}
  \caption{Properties of test example groups, 
  where grouping is based on NoFT prediction outcome: (1) Top-K succeeds; (2) Cover-LS succeeds; (3) only Cover-LS succeeds; and (4) both fail.
  }
  \label{fig:instance}
\end{figure}

\tightparagraph{Stratified analysis}
Our main results show that Cover-LS outperforms Top-K in most compositional splits. But what examples does it perform better on? We analyze properties of test example groups, where grouping is based on NoFT prediction outcome: (1) Top-K succeeds; (2) Cover-LS succeeds; (3) only Cover-LS succeeds; and (4) both fail.
For each group we estimate difficulty by measuring the average accuracy achieved by a T5 model (finetuned without prompts), and also compute the percentage of examples that have an \emph{unobserved local structure} (ULS) with respect to the training set. This measure is central to determining whether generalization to a test instance is hard, as shown in \citet{bogin-etal-2022-unobserved}.\footnote{To comply with \citet{bogin-etal-2022-unobserved}, we measure ULS only for structures up to size 4.}

We see (Fig. \ref{fig:instance}) that as the group index increases, T5 accuracy decreases and ULS rate increases.
This finding confirms the claim in \citet{bogin-etal-2022-unobserved} that a test instance containing an ULS is hard. Examining groups 1 and 3, we observe that the group for which Cover-LS performs better than Top-K, is also tougher for T5 and has more ULS. Both methods fail on examples with low T5 accuracy and high ULS scores (group 4). 
This is also an evidence that T5 and Codex agree on the difficulty of examples, despite their different training and inference schemes. We provide error analysis in App. \ref{app:additional_analysis}.

\tightparagraph{Prompt metrics} 
We analyze the characteristics of prompts constructed with different demonstration selection methods in Tab.~\ref{tab:prompt_metrics}. Symbol Coverage shows the average fraction of symbols in $y_{\small\text{test}}$ that are covered by the demonstration set, and similarly LS Coverage the fraction of covered LSs. 
While symbol coverage is generally high across all methods when using 24 demonstrations, LS coverage is significantly higher in Cover-LS, suggesting that only covering relevant symbols in prompts isn't as efficient as covering LSs.
Utterance Similarity measures average cosine similarity between SBERT embeddings of the test utterance and prompt utterances, which is highest for Top-K as expected.
To approximate diversity between demonstrations, we calculate the average number of unique LSs in demonstrations, and observe it is substantially higher in Cover-LS and DPP compared to Top-K.
This implies structural coverage and diversity are more important than input similarity in compositional splits.

\input{tables/table_prompt_analysis}

\tightparagraph{Robustness to retrieval methods}\label{sec:retrivers} To assess our method's robustness, we test how sensitive it is to the chosen retriever in the NoFT setup. First, we use our default retrievers, which are BM25 over utterance words (BM25-Utterance), and BM25 over predicted program symbols (BM25-Predicted). We add a random retriever that is identical to the \textsc{Random} baseline introduced in \S\ref{sec:baselines} when combined with Top-K. We also evaluate the SBERT retriever \citep{reimers-gurevych-2019-sentence}, which encodes input utterances and measures the cosine similarity between pairs of encodings. As seen in Fig. \ref{fig:retrivers}, Cover-LS outperforms Top-K in all settings by a significant margin.
Moreover, while BM25-Utterance performs best, variance across retrievers is low for Cover-LS, but higher for Top-K.

%% file: tables/table_NFT.tex
\begin{table*}[!ht]
    \centering
    \footnotesize
    
    \begin{threeparttable}
\begin{tabular}{@{}lcccccccccc@{}}
\toprule
 & \multicolumn{4}{c}{\textbf{GeoQuery}} & \multicolumn{5}{c}{\textbf{SMCalFlow-CS}} & \textbf{COVR-10} \\
 \cmidrule[0.4pt](r{0.125em}){2-5}%
 \cmidrule[0.4pt](lr{0.125em}){6-10}%
 \cmidrule[0.4pt](lr{0.125em}){11-11}%
 &  i.i.d. & Templ. & TMCD & Len. & i.i.d. & 0-C & 8-C & 16-C & 32-C &  \\ 
\midrule
T5 (fine tuned w/o prompts) & 90.3 & 85.9 & 75.4 & 36.0 & 88.5 & 0.0 & 34.5 & 39.0 & 50.0 & 21.5 \\
\midrule
Random & 53.7 & 49.7 & 42.0 & 30.7 & 43.0 & 1.3  & 0.3  & 0.7  & 2.0  & 69.4 \\
Top-K & 86.3 & 78.0 & 71.8 & 64.3 & 81.7 & 17.0 & 34.0 & 35.7 & 50.3 & 61.8 \\
Cover-Utt (ours) & \textbf{89.0} & 82.1 & 77.8 & 73.7 & 83.3 & \textbf{35.3} & 51.0 & 51.3 & 69.7 & \textbf{78.1} \\
DPP (ours) & 87.0 & 81.2 & 77.8 & \textbf{74.3} & 79.3 & 34.7 & 44.0 & 50.0 & 59.7 & 62.7 \\
Cover-LS (ours) & 88.7 & \textbf{85.3} & \textbf{79.4} & 72.7 & \textbf{86.0} & 0.3 & \textbf{53.3} & \textbf{58.3} & \textbf{73.5} & 64.4 \\
\midrule
Top-K (Oracle) & 86.3 & 74.5 & 76.2 & 55.7 & 85.0 & 0.0 & 33.0 & 54.0 & 59.6 & 35.4 \\
Cover-LS (Oracle) & 86.3 & 81.2 & 82.8 & 74.0     & 84.3 & 40.7 & 77.3 & 73.5 & 75.3 & 83.2 \\ 
\bottomrule
\end{tabular}
\end{threeparttable}%

\caption{\textbf{Main results, NoFT setup.} We show results of the Codex model on a random subset of 100 test examples across 3 seeds, with the results of a finetuned T5 model for comparison.
}
\label{tab:NFT}
\end{table*}

%% file: tables/table_entire_test_alternative.tex
\newcommand{\equalWidthNum}[1]{\parbox{0.5cm}{\centering #1}}

\begin{table*}[!ht]
    \centering
    \scriptsize
    \resizebox{\textwidth}{!}{%
    \begin{threeparttable}
\begin{tabular}{@{}lccccccccc@{}}
\toprule
 & \multicolumn{4}{c}{\textbf{GeoQuery}} & \multicolumn{5}{c}{\textbf{SMCalFlow-CS}} \\
 \cmidrule[0.4pt](r{0.125em}){2-5}%
 \cmidrule[0.4pt](lr{0.125em}){6-10}%
 &  i.i.d. & Templ. & TMCD & Len. & i.i.d. & 0-C & 8-C & 16-C & 32-C \\ 
 \midrule
T5 Base (FT, $\textbf{\citealt{qiu-etal-2022-improving}}$) & 93.3 & 84.8 & 69.2 & 41.8 & \equalWidthNum{84.7} / \equalWidthNum{-} & - & \equalWidthNum{34.7} / \equalWidthNum{-} & \equalWidthNum{44.7} / \equalWidthNum{-} & \equalWidthNum{59.0} / \equalWidthNum{-} \\
T5 Base + CSL-Aug (FT, $\textbf{\citealt{qiu-etal-2022-improving}}$) & 93.3 & \textbf{89.3} & 74.9 & 67.8 & \equalWidthNum{83.5} / \equalWidthNum{-} & \equalWidthNum{-} & \equalWidthNum{51.6} / \equalWidthNum{-} & \equalWidthNum{\textbf{61.4}} / \equalWidthNum{-} & \equalWidthNum{70.4} / \equalWidthNum{-} \\
T5 Base (FT, $\textbf{\citealt{qiu-etal-2022-evaluating}}$) & 92.9 & 84.8 & 69.2 & 40.0 & \equalWidthNum{-} / \equalWidthNum{82.8} & \equalWidthNum{-} & \equalWidthNum{-} / \equalWidthNum{21.7} & \equalWidthNum{-} / \equalWidthNum{43.6} & \equalWidthNum{-} / \equalWidthNum{58.9} \\
T5 11B (Prompt Tuning, $\textbf{\citealt{qiu-etal-2022-evaluating}}$) & \textbf{93.6} & 87.7 & \textbf{81.2} & 41.5 & \equalWidthNum{-} / \equalWidthNum{83.1} & \equalWidthNum{-} & \equalWidthNum{-} / \equalWidthNum{0.0} & \equalWidthNum{-} / \equalWidthNum{10.0} & \equalWidthNum{-} / \equalWidthNum{23.6} \\
PaLM 62B (FT, $\textbf{\citealt{qiu-etal-2022-evaluating}}$) & 92.5 & 85.1 & 72.7 & 44.2 & \equalWidthNum{-} / \equalWidthNum{82.2} & \equalWidthNum{-} & \equalWidthNum{-} / \equalWidthNum{26.9} & \equalWidthNum{-} / \equalWidthNum{34.7} & \equalWidthNum{-} / \equalWidthNum{51.1} \\
PaLM 540B (ICL, $\textbf{\citealt{qiu-etal-2022-evaluating}}$) & 86.8 & 76.6 & 63.6 & 57.9 & \equalWidthNum{-} / \equalWidthNum{58.3} & \equalWidthNum{-} & \equalWidthNum{-} / \equalWidthNum{4.7} & \equalWidthNum{-} / \equalWidthNum{5.0} & \equalWidthNum{-} / \equalWidthNum{11.7}\\

\midrule

T5 Large (fine tuned w/o prompts) & 92.5 & 83.8 & 73.5 & 37.2 & \equalWidthNum{\textbf{85.3}} / \equalWidthNum{\textbf{83.3}}  & \equalWidthNum{0.0} / \equalWidthNum{0.0} & \equalWidthNum{34.3} / \equalWidthNum{6.9}  & \equalWidthNum{43.0} / \equalWidthNum{33.6}  & \equalWidthNum{56.1} / \equalWidthNum{53.6} \\
Top-K (NoFT) & 88.9 & 74.7 & 69.4 & 65.8 & \equalWidthNum{79.3} / \equalWidthNum{69.7} & \equalWidthNum{\textbf{19.8}} / \equalWidthNum{\textbf{13.6}} & \equalWidthNum{32.7} / \equalWidthNum{25.8} & \equalWidthNum{37.7} / \equalWidthNum{33.6} & \equalWidthNum{49.6} / \equalWidthNum{43.9} \\
Cover-LS (NoFT) & 91.4 & 81.6 & 76.3 & \textbf{70.0} & \equalWidthNum{82.2} / \equalWidthNum{73.6}  & \equalWidthNum{0.0} / \equalWidthNum{0.0}  & \textbf{52.5} / \equalWidthNum{\textbf{36.7}}  & \equalWidthNum{60.9} / \equalWidthNum{\textbf{60.3}}  & \equalWidthNum{\textbf{75.1}} / \equalWidthNum{\textbf{64.7}}  \\
\bottomrule
\end{tabular}
\end{threeparttable}%
}
\caption{\textbf{NoFT setup compared to past approaches} on the entire test set (single seed). Since past work reported results on different versions of SMCalFlow-CS, we report accuracy for both versions (v1 / v2).}
\label{tab:entire_test_set}
\end{table*}

%% file: tables/table_FT.tex
\begin{table*}[!ht]
\centering
\footnotesize
\begin{tabular}{@{}llccccccccc@{}}
\toprule
\textbf{Training Method} & \textbf{Test Method} & \multicolumn{4}{c}{\textbf{GeoQuery}} & \multicolumn{4}{c}{\textbf{SMCalFlow-CS Simple}} & \textbf{COVR-10} \\
\cmidrule[0.4pt](r{0.125em}){3-6}%
\cmidrule[0.4pt](lr{0.125em}){7-10}%
\cmidrule[0.4pt](lr{0.125em}){11-11}%
 &  & i.i.d. & Templ. & TMCD & Len. & i.i.d. & 8-C & 16-C & 32-C &  \\
\midrule
T5 (FT, w/o prompts) & - & 92.5 & 83.8 & 73.5 & 37.2 & 83.7 & 9.7 & 37.5 & 59.4 & 19.4 \\ 
\midrule
Random & Random & \textbf{93.2} & 85.0 & 76.8 & 39.8 & 83.5 & 28.3 & 46.4 & 58.0 & 23.2 \\
Random & Top-K & 93.0 & 84.6 & 75.9 & 39.8 & 83.4 & 24.4 & 40.6 & 54.8 & 22.8 \\
Top-K & Top-K & 90.7 & 54.7 & 57.4 & 20.8 & 83.2 & 8.8 & 22.1 & 46.1 & 19.6 \\
$\textrm{Cover-LS}_{1}$ & $\textrm{Cover-LS}_{1}$ & 92.9 & 85.3 & 76.6 & 41.9 & 83.9 & \textbf{31.0} & \textbf{51.3} & \textbf{62.6} & \textbf{29.8} \\
$\textrm{Cover-LS}_{1}$ & Cover-LS & 93.1 & \textbf{85.9} & \textbf{77.6} & \textbf{42.7} & \textbf{84.1} & 30.5 & 50.6 & 61.5 & 28.6  \\
\midrule
$\textrm{Cover-LS}_{2}$ & Cover-LS & 92.6 & 84.9 & 75.6 & 39.8 & 83.7 & 28.8 & 46.3 & 60.5 & 28.8 \\
Cover-LS & Cover-LS & 91.8 & 80.7 & 69.4 & 37.7 & 82.9 & 21.2 & 34.1 & 53.8 & 13.6 \\
\midrule
$\textrm{Cover-LS}_{1}$ & Cover-LS (Oracle) & 93.7 & 87.7 & 79.8 & 48.9 & 87.4 & 48.0 & 64.1 & 73.5 & 41.1 \\ 
\bottomrule
\end{tabular}%
\caption{\textbf{FT results} using T5. We detail the method used for demonstration selection at both training time and test time as those may differ to avoid over-copying.}
\label{tab:FT}
\end{table*}

%% file: tables/table_prompt_analysis.tex
\begin{table}[!t]
\vspace{4pt}
\resizebox{\columnwidth}{!}{%
\begin{tabular}{@{}lcccccc@{}}
\toprule
\textbf{Prompt Metrics} & \multicolumn{3}{c}{\textbf{GeoQuery TMCD}} & \multicolumn{3}{c}{\textbf{SMCalFlow-CS 8-C}} \\
 & Top-K & Cover-LS & DPP & Top-K & Cover-LS & DPP \\
\midrule
Symbol Coverage & 97.2 & 99.3 & 99.2 & 93.1 & 95.0 & 96.6 \\
LS Coverage & 69.2 & 73.0 & 71.0 & 70.0 & 86.8 & 76.1 \\
Utterance Sim. & 0.46 & 0.42 & 0.43 & 0.50 & 0.47 & 0.48 \\ 
No. Unique LSs & 306 & 505 & 484 & 2139 & 3647 & 4212 \\ 
\bottomrule
\end{tabular}%
}
\caption{Prompt metrics: coverage, similarity, and diversity in prompts with 24 demonstrations.}
\label{tab:prompt_metrics}
\end{table}

%% file: 5_related.tex
\section{Related Work}
\tightparagraph{Example selection}
One of the central issues in in-context learning is the selection of examples, which can either be based on parameter-free retrievers \cite{wang-etal-2022-training,zemlyanskiy-etal-2022-generate} or neural-based retrievers \cite{pasupat-etal-2021-controllable,liu-etal-2022-makes,rubin-etal-2022-learning}. These studies consider each example separately, which often leads to a lack of coverage and diversity.

Our approach is similar to the retrieval procedure in \citet{zemlyanskiy-etal-2022-generate}, which makes a preliminary prediction and retrieves demonstrations with similar programs. However, while they use classic tf-idf with predicted tokens, we use predicted local structures and aim to cover them.

Some studies encourage diverse example selection regardless of prompting. To address multi-answer retrieval, \citet{nandigam-etal-2022-diverse} employ DPP, and \citet{min-etal-2021-joint}  autoregressively select instances based on previous selections. 
Other works include \citet{Su2022SelectiveAM}, which selects instances with varying confidence scores for annotation 
and (concurrent work) \citet{Ye2022ComplementaryEF} who propose 
a maximum marginal relevance-based selection strategy.

\tightparagraph{In-context learning for compositional generalization}
There have been previous attempts to address compositional generalization problems using LLMs equipped with demonstrations. When selecting demonstrations, some also consider target coverage or structure similarity, but only in oracle setups \cite{hosseini-etal-2022-compositional,qiu-etal-2022-evaluating}. \citet{Drozdov2022CompositionalSP} try to cover the syntactic parse tree constituents with demonstrations but rely heavily on manually-picked examples.

\begin{figure}[!t]
  \centering
  \includegraphics[width=\linewidth]{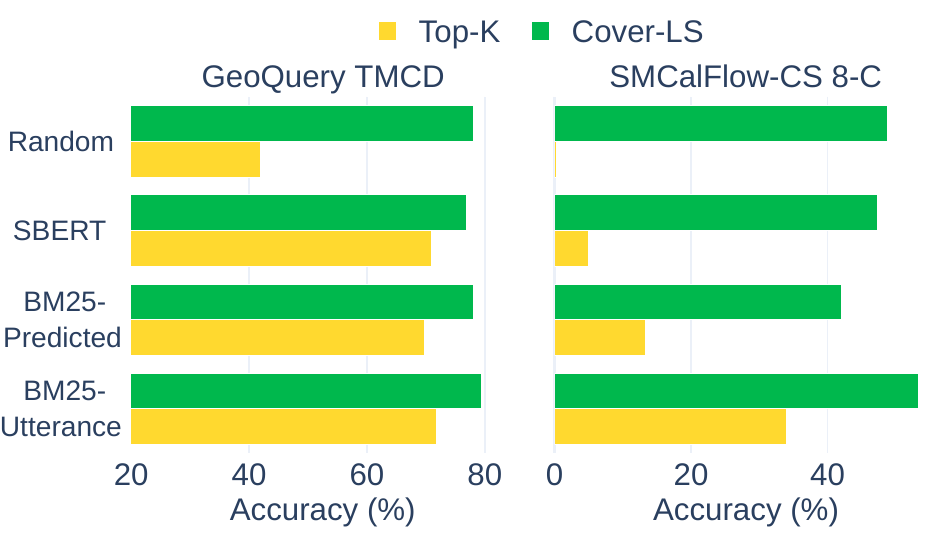}
  \caption{Comparing model accuracy across different retrievers, with demonstrations selected using Top-K or Cover-LS.
  }
  \label{fig:retrivers}
\end{figure}

%% file: 6_conclusion.tex
\section{Conclusion}
In this paper, we studied how to leverage ICL to improve compositional generalization in semantic parsing, by increasing diversity among demonstrations. We found that choosing demonstrations that cover the structures required in the output program substantially improves performance across three compositional semantic parsing datasets in the pure in-context learning setup and when combined with finetuning. We further demonstrated that by aiming for structural coverage, we can reduce the number of demonstrations needed for generalization, and improve test performance on hard examples. Our approach can be applied to a wide range of NLP tasks where demonstrations should cover complementary aspects of the task, and we hope it will encourage further exploration of our method to improve generalization across diverse applications.

%% file: 7_limitations.tex
\section*{Limitations}

\tightparagraph{Demonstration selection methods}
We assume that diversity can be obtained by choosing demonstrations with different program structures. This is based on previous work that demonstrated the importance of diversifying program structures in semantic parsing tasks \cite{oren-etal-2021-finding,bogin-etal-2022-unobserved,gupta-etal-2022-structurally}. We also try to diversify utterance words or program symbols but do not consider more complex utterance features that could be applied to a wider range of language understating tasks.

We also assume that recall matters more than precision when designing Cover-LS algorithm. That means we aim to choose a set of demonstrations that covers every predicted local structure in $\mathcal{S}_{\tilde{y}_{\small\text{test}}}$, since it has the potential to be a correct one. We do not predict whether a specific structure should be covered. Furthermore, our approach for increasing gold structure coverage by using additional beam candidates could be improved by employing search methods specifically targeted for diversity \cite{meister-etal-2021-determinantal,narayan-etal-2022-well}.

\tightparagraph{Retrievers}
We used different retrievers for NoFT and FT setups based on the retriever that worked best on the development set. Future research should be conducted to understand why different retrievers are preferred in different setups. A potential method could be to consider both input utterances and programs for retrieval, as suggested in \citet{zemlyanskiy-etal-2022-generate}.

\section*{Ethics Statement}
In this work, we studied methods for choosing diverse demonstrations to improve in-context compositional generalization in semantic parsing. We have only evaluated our methods on semantic parsing datasets in English. It is our hope, however, that improvements in compositional generalization will eventually allow systems to generalize better to languages that are not well represented in small training sets.

%% file: 8_acknowledgments.tex
\section*{Acknowledgements}
We thank Shivanshu Gupta and Jonathan Herzig for their helpful comments. This research was partially supported by The Yandex Initiative for Machine Learning, and the European Research Council (ERC) under the European Union Horizons 2020 research and innovation programme (grant ERC DELPHI 802800). This work was completed in partial fulfillment for the Ph.D degree of Ben Bogin.

%% file: app_a.tex
\section{Additional Analysis}
\label{app:additional_analysis}

\tightparagraph{Error analysis} We analyze errors (NoFT setup) and show results in Tab.~\ref{tab:error}. Inspired by the metrics in \citet{qiu-etal-2022-evaluating}, we automatically compute statistics for the following cases when the prediction is wrong: (1) Syntax Errors, when the model produces a program with invalid parentheses; (2) Over-Copying, when the entire prediction has the same anonymized form as one of the demonstrations; (3) OOV (out-of-vocabulary) Hallucination, where the anonymized predicted program contains a symbol missing from the gold program or any prompt demonstration; and (4) Missing Symbol(s), where the predicted program is missing at least one symbol. 

The distribution of errors is similar across demonstration selection methods. Syntax errors are rare in both datasets. Many predictions are over-copied, especially in SMCalFlow-CS, but when diversity is increased with DPP, this number decreases significantly. Surprisingly, despite having a smaller vocabulary, GeoQuery has more out-of-vocabulary hallucinations. Almost all incorrect predictions have a missing symbol, but Top-K predictions are especially prone to this type of error.

\tightparagraph{Change of retriever in FT setup}
Tab.~\ref{tab:FT_source} shows results for the FT setup when using BM25 over lower-cased utterance words as retriever, instead of BM25 over predicted program symbols.

\section{Local Structures}
\label{app:local_structures}

We follow the definition of local structures from \citet{bogin-etal-2022-unobserved}, which were defined for structures of sizes 2-4, and extend them to local structures of any size. Given a program $y$, we parse it into a tree $T=(\mathcal{V},\mathcal{E})$, such that each node $v \in \mathcal{V}$ is labeled by the program symbol (function or value) that it represents in $y$ (or a special symbol for the root node),
and the set of edges $\mathcal{E}=\{(p,c)\}$ expresses \textbf{parent-child} relations between the nodes.

We capture sibling relations by defining a graph based on the tree $T$ that contains an 
edge set $\mathcal{E}_{\text{sib}}$ of \textbf{sibling} edges: 
$G=(\mathcal{V},\mathcal{E} \cup \mathcal{E}_{\text{sib}})$. 
Specifically, for each parent node $p$, the program $y$ induces an order over the children of $p$: $(c^p_1, ..., c^p_{N_p})$, where $N_p$ is the number of children. We then define $\mathcal{E}_{\text{sib}}=\bigcup_p\{c_i^p, c_{i+1}^p\}_{i=1}^{N_p}$, that is, all \emph{consecutive} siblings will be connected by edges.

We define a local structure of size $n$ as the subset $G_{LS}$ of all connected sub-graphs of size $n$ in $G$ such that for every pair $(x, y)$ of nodes in $G_{LS}$ it holds that $(x, y) \in \mathcal{E}_{\text{sib}}$ iff $x$ and $y$ are both leaves in $G_{LS}$. That is, informally, the relations between nodes in the the sub-graph include parent-child and siblings, but not e.g. cousins or uncles. All program symbols are local structures of size 1. Tab.~\ref{tab:ls_examples} shows a partial list of local structures for a given program.

\input{tables/table_error_analysis}
\input{tables/app/table_FT_source}

\subsection{Fixes for Local Structure Extraction}
We try to fix syntax errors in the predictions made using the auxiliary model to enable parsing them to ASTs and extraction of LSs. We add or remove closing parentheses based on the number of missing or redundant parentheses at the end of the program.

\section{Dataset Details}\label{app:sizes}
We provide representative examples of the datasets used in this work in Tab.~\ref{tab:datasets} and Tab.~\ref{tab:app_smcal}. We report dataset sizes in Tab.~\ref{tab:dataset_sizes}.
Due to conversion errors, SMCalFlow-CS Simple has fewer training examples than SMCalFlow-CS. However, those missing examples are not cross-domain examples. 

We used publicly available datasets from previous peer-reviewed studies. Those datasets do not contain any information that uniquely identifies individual people or offensive content. The COVR-10 dataset is completely synthetic. The GeoQuery dataset contains only basic information about U.S. geography. SMCalflow-CS contains crowd-sourced queries collected in a simulated environment. 

\section{Prompt Format and Examples}\label{app:prompt_format}
We add special prefixes ``source:'' and ``target:'' for retrieved source-target pairs and separate them with break lines. Tab.~\ref{tab:prompt_examples} shows prompt examples for different demonstration selection methods, where the only prompt that contains all the required program symbols and produces the correct prediction is Cover-LS's prompt.

\section{DPP Details}\label{app:dpp}
DPPs are probabilistic models that
are effective at modeling a distribution on all the subsets of the ground set $\mathcal{T}$ jointly considering the quality and diversity.
A subset $\mathcal{D}$ is drawn
according to the probability distribution $\mathcal{P}$:

\begin{equation}
\mathcal{P}(\mathcal{D} \subset \mathcal{T}; L) \propto \det(L_\mathcal{D}) 
\end{equation}

\noindent 
Where $L \in \mathbb{R}^{n \times n}$ is a PSD matrix and $L_\mathcal{D}$
is the submatrix of $L$ indexed by items in $\mathcal{D}$. $L$ matrix takes into account the quality of each training example and its similarity to other training examples through:

\begin{equation}
L_{ij}=q_i \phi_i^\top \phi_j q_j
\end{equation}
with $q \in \mathbb{R}^{n}$ being normalized retriever scores that model the quality of each example; and $\{\phi_i\}_{i=1}^n$ denoting normalized tf-idf vectors over LSs, which model the different aspects that are contained within each training example. The dot product of those vectors is used to model the similarity between two train examples. 

$\log \ \det(L_\mathcal{D})$ is a submodular function which satisfies the diminishing marginal returns property. Therefore, we can find a subset of training examples $\mathcal{D} \subset \mathcal{T}, |\mathcal{D}|=k$ that maximizes it in a feasible manner using a greedy optimizer \citep{Kaushal2022SubmodlibAS}. Specifically, we used the Naive Greedy optimizer.
We used scikit-learn \cite{scikit-learn} for calculating tf-idf vectors.

\section{Finetuning Details}
\label{app:FT}
We provide implementation details for finetuning experiments (we use the same configuration for all FT experiments and training of the auxiliary model). We finetune the T5-large model (770 million parameters) with the AdamW optimizer \cite{Loshchilov2017DecoupledWD} and a learning rate of $1e^{-5}$. We use a polynomial decay learning rate with an ending rate of $1e^{-6}$, and 100 warmup steps. We train for 250/50/70 epochs and evaluate on the validation set every 3/5/10 epochs for Geo/SMCalFlow (both versions)/COVR respectively. We use batches of size 8 for all datasets (and gradient accumulation in case batch cannot fit in memory). We used a single GPU for each T5-large finetuning experiment: Nvidia GeForce RTX 3090 when training on GeoQuery and COVR-10, and A100 (80GB) for SMCalFlow-CS and SMCalFlow-CS Simple. GeoQuery experiments with prompts trained for an average of 2 hours, COVR for 8 hours, and SMCalFlow-CS Simple for 41 hours.

We use the AllenNLP library \citep{gardner-etal-2018-allennlp} for training and evaluation. 
We use Rank-BM25 \cite{rank_bm25} as a BM25 implementation.

\paragraph{Standard deviation}\label{app:additional_FT_results}
We report standard deviation results in the FT setup in Tab.~\ref{tab:std_ft}. Results are computed across 3 random seeds.

\section{NoFT Details}
All NoFT experiments were conducted using the OpenAI inference API with the sampling temperature set to 0. Our setup requires a single API call per test instance. The total number of API calls is estimated at 160K.

\paragraph{Standard deviation}\label{app:additional_NFT_results}
We report standard deviation results in NoFT setup in Tab.~\ref{tab:std_nft}. Results are computed using 3 random seeds for a subset of 100 test examples.

\paragraph{Tuning the number of beam candidates}
We use the development set to tune the number of beam
candidates $B$ when predicting local structures. Tab.~\ref{tab:beam_size} shows the results of using different values of $B$ in NoFT setup on a random subset of 100 development examples. Prompts are constructed using
Cover-LS with k = 8 demonstrations. 

\section{Artifact Licensing}
We include license information for all artifacts used in this work in Tab.~\ref{tab:license}. Our use of artifacts was consistent with their intended purpose when it was specified.

\section{GenBench Evaluation Card}
\label{app:gen_bench}
\begin{figure*}[!t]
  \centering
  \includegraphics[width=0.9\textwidth]{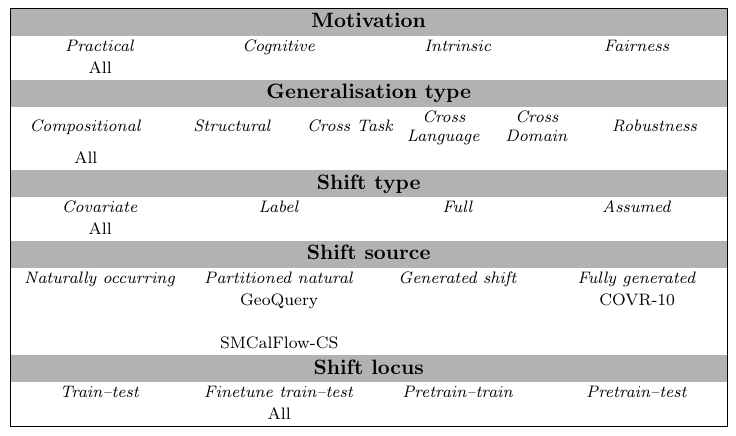}
  \setlength{\belowcaptionskip}{-5pt}
  \caption{GenBench \cite{hupkes2022taxonomy} evaluation card. 
  }
  \label{fig:genbench}
\end{figure*}

Our GenBench \cite{hupkes2022taxonomy} evaluation card is presented in Fig.~\ref{fig:genbench}.

\input{tables/app/ls_example}
\input{tables/app/table_smcal_versions}
\input{tables/app/dataset_sizes}
\input{tables/app/prompt_examples}
\input{tables/app/std_nft}
\input{tables/app/std_ft}
\input{tables/app/beam_size_nft}
\input{tables/app/license_artifacts.tex}

%% file: tables/table_error_analysis.tex
\begin{table}[!t]
\vspace{4pt}
\resizebox{\columnwidth}{!}{%
\begin{tabular}{@{}lcccccc@{}}
\toprule
\textbf{Error Types} & \multicolumn{3}{c}{\textbf{GeoQuery TMCD}} & \multicolumn{3}{c}{\textbf{SMCalFlow-CS 8-C}} \\
 & \multicolumn{1}{c}{Top-K} & \multicolumn{1}{c}{Cover-LS} & \multicolumn{1}{c}{DPP} & \multicolumn{1}{c}{Top-K} & \multicolumn{1}{c}{Cover-LS} & \multicolumn{1}{c}{DPP} \\
\midrule
Syntax Error & 1.0 & 0.0 & 0.9 & 5.0 & 2.9 & 9.5 \\
Over-Copying & 19.8 & 16.9 & 15.8 & 41.4 & 41.4 & 10.7 \\
OOV Hallucination & 20.0 & 17.8 & 22.9 & 8.0 & 3.5 & 5.4 \\
Missing Symbol(s) & 88.7 & 75.2 & 77.9 & 87.4 & 77.7 & 79.8 \\
\bottomrule
\end{tabular}%
}
\caption{Error analysis. We automatically compute the fraction of different error types.}
\label{tab:error}
\end{table}

%% file: tables/app/table_FT_source.tex
\begin{table*}[!ht]
\centering
\small
\begin{tabular}{@{}llccccccccc@{}}
\toprule
\textbf{Training Method} & \textbf{Test Method} & \multicolumn{4}{c}{\textbf{GeoQuery}} & \multicolumn{4}{c}{\textbf{SMCalFlow-CS Simple}} & \textbf{COVR-10} \\
\cmidrule[0.4pt](r{0.125em}){3-6}%
\cmidrule[0.4pt](lr{0.125em}){7-10}%
\cmidrule[0.4pt](lr{0.125em}){11-11}%
 &  & i.i.d. & Templ. & TMCD & Len. & i.i.d. & 8-C & 16-C & 32-C &  \\
\midrule
Random & Top-K & 93.0 & 84.9 & 76.1 & 40.3 & 82.9 & 26.7 & 41.0 & 53.9 & 23.1 \\
$\textrm{Cover-LS}_{1}$ & $\textrm{Cover-LS}_{1}$ & \textbf{93.3} & 85.7 & 76.3 & 42.2 & \textbf{83.2} & \textbf{31.9} & \textbf{48.6} & \textbf{61.5} & 28.3 \\
$\textrm{Cover-LS}_{1}$ & Cover-LS & 93.2 & \textbf{85.8} & \textbf{76.6} & \textbf{42.4} & \textbf{83.2} & 28.3 & 46.6 & 60.9 & \textbf{30.1}  \\
\midrule
$\textrm{Cover-LS}_{2}$ & Cover-LS & 92.5 & 85.2 & 75.1 & 39.7 & 83.9 & 27.2 & 45.5 & 59.5 & 29.8 \\
Cover-LS & Cover-LS & 91.4 & 81.0 & 69.1 & 39.2 & 82.7 & 17.5 & 31.5 & 55.1 & 12.3 \\
\bottomrule
\end{tabular}%
\caption{\textbf{FT results} using T5. Same setup as in Tab.~\ref{tab:FT}, except we use BM25 over lower-cased utterance words.}
\label{tab:FT_source}
\end{table*}

%% file: tables/app/ls_example.tex
\begin{table*}[!ht]
\centering
\scriptsize
\begin{tabular}{p{2.55cm}p{12.1cm}}
\toprule
\textbf{Dataset} & SMCalFlow-CS Simple \\
\textbf{Utterance} & \textit{Create a new meeting on Friday called Work on Project.} \\
\textbf{Program} & \texttt{CreateEvent (AND (has\_subject (``Work on Project''), starts\_at (NextDOW (``Friday''))))} \\

\textbf{Anonymized Program} & \texttt{CreateEvent (AND (has\_subject (string), starts\_at (NextDOW (string))))} \\
\toprule
{\bf Size} & {\bf Local structures} \\
 \midrule
 \multirow{6}{*}{1} & 
 \texttt{CreateEvent } \\ 
 & \texttt{AND } \\
 & \texttt{has\_subject} \\ 
 & \texttt{string} \\  
 & \texttt{starts\_at} \\ 
 & \texttt{NextDOW } \\
 \midrule
 \multirow{8}{*}{2} & 
  \texttt{<root> $\rightarrow$ CreateEvent} \\
  & \texttt{CreateEvent $\rightarrow$ AND} \\
  & \texttt{AND $\rightarrow$ has\_subject} \\
  & \texttt{AND $\rightarrow$ starts\_at} \\
  & \texttt{has\_subject $\leftrightarrow$ starts\_at} \\
  & \texttt{has\_subject $\rightarrow$ string} \\
  & \texttt{starts\_at $\rightarrow$ NextDOW} \\
  & \texttt{NextDOW $\rightarrow$ string}
  \\
  \midrule
  \multirow{7}{*}{3} & 
  \texttt{<root> $\rightarrow$ CreateEvent $\rightarrow$ AND} \\
  & \texttt{CreateEvent $\rightarrow$ AND $\rightarrow$ has\_subject} \\
  & \texttt{CreateEvent $\rightarrow$ AND $\rightarrow$ starts\_at} \\
  & \texttt{AND $\rightarrow$ has\_subject $\leftrightarrow$ starts\_at} \\
  & \texttt{AND $\rightarrow$ has\_subject $\rightarrow$ string} \\
  & \texttt{AND $\rightarrow$ starts\_at $\rightarrow$ NextDOW} \\
  & \texttt{starts\_at $\rightarrow$ NextDOW $\rightarrow$ string}\\
  \midrule
  & \vdots \\
  \midrule
  6 & \texttt{<root> $\rightarrow$ CreateEvent $\rightarrow$ AND $\rightarrow$ starts\_at $\rightarrow$ NextDOW $\rightarrow$ string} \\
 \bottomrule
\end{tabular}
\caption{Local structures of different sizes for a specific example ($\rightarrow$ denotes parent-child relations, $\leftrightarrow$ denotes sibling relations)}
\label{tab:ls_examples}
\end{table*}

%% file: tables/app/table_smcal_versions.tex
\begin{table*}[!t]

\centering
    \scriptsize
    \resizebox{\textwidth}{!}{%
    \begin{threeparttable}
\begin{tabular}{@{}ll@{}}

\toprule
\textbf{Utterance} & \textit{Can you make a meeting with David Lax 's reports ?}\\ 
\toprule
{\bf Version} & {\bf Program} \\
 \midrule
  \makecell[l]{\textbf{v1}
  } (LISP) & \makecell[l]{
  \texttt{(Yield :output (CreateCommitEventWrapper :event (CreatePreflightEventWrapper} \\
  \texttt{:constraint (Constraint[Event] :attendees (AttendeeListHasPeople}  
  \\\texttt{:people (FindReports :recipient (Execute :intension (refer (extensionConstraint} \\ \texttt{(RecipientWithNameLike :constraint (Constraint[Recipient]) :name } \\ \texttt{\# (PersonName ``David Lax")))))))))))}}\\ 

\cmidrule[0.4pt](r{0.125em}){1-2}%
  \makecell[l]{\textbf{v2} (LISPRESS)} 
  & \makecell[l]{\texttt{(Yield (CreateCommitEventWrapper (CreatePreflightEventWrapper (Event.attendees\_?} \\
  \texttt{(AttendeeListHasPeople (FindReports (Execute (refer (extensionConstraint} \\ \texttt{(RecipientWithNameLike ( \string^ (Recipient) EmptyStructConstraint)} \\
  \texttt{(PersonName.apply ``David Lax")))))))))))}
  }\\

 \bottomrule
\end{tabular}
\end{threeparttable}%
}
\caption{An example from each version of SMCalFlow-CS dataset.
}
\label{tab:app_smcal}
\end{table*}

%% file: tables/app/dataset_sizes.tex
\begin{table*}[!ht]
\centering
\small
\begin{tabular}{@{}llccc@{}}
\toprule
Dataset & Split & \multicolumn{1}{c}{Train} & \multicolumn{1}{c}{Development} & \multicolumn{1}{c}{Test} \\ \midrule
\multirow{8}{*}{GeoQuery} & Standard & 600 & - & 280 \\
 & Template1 & 438 & 110 & 332 \\
 & Template2 & 439 & 110 & 331 \\
 & Template3 & 440 & 110 & 330 \\
 & TMCD1 & 440 & 110 & 330 \\
 & TMCD2 & 440 & 110 & 330 \\
 & TMCD3 & 440 & 110 & 330 \\
 & Length & 440 & 110 & 330 \\ \midrule
\multirow{5}{*}{SMCalFlow-CS v1} & 8-S & 25412 & 662 & 662 \\
 & 0-C & 25404 & 662 & 663 \\
 & 8-C & 25412 & 662 & 663 \\
 & 16-C & 25420 & 662 & 663 \\
 & 32-C & 25436 & 662 & 663 \\ \midrule 
 \multirow{5}{*}{SMCalFlow-CS v2} & 8-S & 20965 & 360 & 360 \\
 & 0-C & 20957 & 360 & 360 \\
 & 8-C & 20965 & 360 & 360 \\
 & 16-C & 20973 & 360 & 360 \\
 & 32-C & 20989 & 360 & 360 \\ \midrule
\multirow{4}{*}{\begin{tabular}[c]{@{}l@{}}SMCalFlow-CS \\ Simple\end{tabular}} & 8-S & 25402 & 662 & 662 \\
 & 8-C & 25402 & 662 & 663 \\
 & 16-C & 25410 & 662 & 663 \\
 & 32-C & 25426 & 662 & 662 \\ \midrule
COVR-10 & Each split & 3000 & - & 500 \\ \bottomrule
\end{tabular}%
\caption{Dataset sizes}
\label{tab:dataset_sizes}
\end{table*}

%% file: tables/app/prompt_examples.tex
\begin{table*}[!ht]
\centering
\small
\begin{tabular}{p{2.55cm}p{12.1cm}}
\toprule
\textbf{Dataset} & GeoQuery \\
\textbf{Utterance} & \textit{through which states does the longest river \colorbox{green}{in texas} run} \\
\textbf{Gold Program} & \texttt{\colorbox{yellow}{answer (state (traverse\_1 (longest (river} \colorbox{green}{(loc\_2 (stateid (string)))))))}} \\
\toprule
{\bf Selection Method} & {\bf Prompt} \\
 \midrule
 \multirow{10}{*}{\bf Top-K} &source: \textit{which states does the mississippi river run through}  \\ 
&target: \colorbox{pink}{\texttt{answer (state (traverse\_1 (river (riverid (string)))))}}\\
&source: \textit{which states does the colorado river run through} \\
&target: \colorbox{pink}{\texttt{answer (state (traverse\_1 (river (riverid (string)))))}}\\
&source: \textit{which states does the missouri river run through} \\
&target: \colorbox{pink}{\texttt{answer (state (traverse\_1 (river (riverid (string)))))}}\\
&source: \textit{which states does the longest river run through} \\
&target: \texttt{\colorbox{yellow}{answer (state (traverse\_1 (longest (river} (all)))))}\\
&source: \textit{through which states does the longest river in texas run} \\
&target:\\
 \midrule
 \multirow{10}{*}{\bf DPP} &source: \textit{what states does the shortest river run through} \\ 
&target: \texttt{answer (state (traverse\_1 (shortest (river (all)))))}\\
&source: \textit{which states does the mississippi run through} \\
&target: \texttt{answer (state (traverse\_1 (riverid (string)))))}\\
&source: \textit{which states does the missouri river run through} \\
&target: \colorbox{pink}{\texttt{answer (state (traverse\_1 (river (riverid (string)))))}}\\
&source: \textit{which states does the longest river run through} \\
&target: \texttt{\colorbox{yellow}{answer (state (traverse\_1 (longest (river} (all)))))}\\
&source: \textit{through which states does the longest river in texas run} \\
&target:\\
\midrule
 \multirow{11}{*}{\bf Cover-LS} &source: \textit{what state borders the least states excluding alaska and excluding hawaii} \\ 
&target: \texttt{answer (fewest (state (next\_to\_2 (exclude (exclude (state (all),}\\
&\texttt{stateid (string)), stateid (string))))))}\\
&source: \textit{what is the longest river \colorbox{green}{in texas}} \\
&target: \texttt{answer (longest (river \colorbox{green}{(loc\_2 (stateid (string)))))}}\\
&source: \textit{which states does the missouri river run through} \\
&target: \colorbox{pink}{\texttt{answer (state (traverse\_1 (river (riverid (string)))))}}\\
&source: \textit{which states does the longest river run through} \\
&target: \texttt{\colorbox{yellow}{answer (state (traverse\_1 (longest (river} (all)))))}\\
&source: \textit{through which states does the longest river in texas run} \\
&target:\\
 
 \bottomrule
\end{tabular}
\caption{Prompts produced with different demonstration selection methods for a specific test example. Each prompt contains $k=4$ demonstrations.}
\vspace{3cm}
\label{tab:prompt_examples}
\end{table*}

%% file: tables/app/std_nft.tex
\begin{table*}[!ht]
    \centering
    \small
    \begin{threeparttable}
\begin{tabular}{@{}lcccccccccc@{}}
\toprule
 & \multicolumn{4}{c}{\textbf{GeoQuery}} & \multicolumn{5}{c}{\textbf{SMCalFlow-CS}} & \textbf{COVR-10} \\
 \cmidrule[0.4pt](r{0.125em}){2-5}%
 \cmidrule[0.4pt](lr{0.125em}){6-10}%
 \cmidrule[0.4pt](lr{0.125em}){11-11}%
 &  i.i.d. & Templ. & TMCD & Len. & i.i.d. & 0-C & 8-C & 16-C & 32-C &  \\ 
\midrule
Random & 1.5	 &  6.6 &  2.5 & 5.0	 & 4.6	 & 0.6	 & 0.6	 & 0.6	 & 3.5	 &  3.1 \\
Top-K & 1.5 & 1.8 & 1.0 & 1.1 & 0.6 & 1.0 & 1.0 & 1.1 & 1.1 & 4.6 \\
Cover-Utt & 1.0 & 1.2 & 1.2 & 2.1 & 1.5 & 1.5 & 1.0 & 1.2 & 2.1 & 1.9 \\
DPP & 0.0 & 0.5 & 1.7 & 1.5 & 1.2 & 0.6 & 1.0 & 1.0 & 3.1 & 2.0 \\
Cover-LS & 1.5 & 1.1 & 2.4 & 2.1 & 1.4 & 0.6 & 1.1 & 0.6 & 3.5 & 4.2\\
\bottomrule
\end{tabular}
\end{threeparttable}%
\caption{Standard deviation results in NoFT setup. Results are computed on a random subset of 100 test examples across 3 random seeds.}
\label{tab:std_nft}
\end{table*}

%% file: tables/app/std_ft.tex
\begin{table*}[!ht]
\centering
\scriptsize

\begin{tabular}{@{}llccccccccc@{}}
\toprule
\textbf{Training Method} & \textbf{Test Method} & \multicolumn{4}{c}{\textbf{GeoQuery}} & \multicolumn{4}{c}{\textbf{SMCalFlow-CS Simple}} & \textbf{COVR-10} \\
\cmidrule[0.4pt](r{0.125em}){3-6}%
\cmidrule[0.4pt](lr{0.125em}){7-10}%
\cmidrule[0.4pt](lr{0.125em}){11-11}%
 &  & i.i.d. & Templ. & TMCD & Len. & i.i.d. & 8-C & 16-C & 32-C &  \\
\midrule
T5 (fine tuned w/o prompts) & - & 0.2 & 0.8 & 1.6 & 0.5 & 0.7 & 1.4 & 4.6 & 1.5 & 1.7 \\ 
\midrule
Random & Random & 0.0 & 1.2 & 1.0 & 0.9 & 0.3 & 3.2 & 2.7 & 0.4 & 2.7 \\
Random & Top-K & 0.2 & 1.4 & 1.3 & 2.3 & 0.4 & 3.3 & 1.2 & 1.2 & 2.7 \\
Top-K & Top-K & 0.6 & 3.5 & 2.1 & 0.7 & 0.3 & 1.9 & 1.9 & 1.3 & 3.9 \\
$\textrm{Cover-LS}_{1}$ & $\textrm{Cover-LS}_{1}$ & 0.6 & 0.8 & 0.9 & 2.6 & 0.5 & 2.0 & 0.2 & 1.7 & 4.8 \\
$\textrm{Cover-LS}_{1}$ & Cover-LS & 0.5 & 0.4 & 0.9 & 4.2 & 0.4 & 1.4 & 0.8 & 0.8 & 6.5 \\
\midrule
$\textrm{Cover-LS}_{1}$ & Cover-LS (Oracle) & 0.2 & 0.7 & 0.9 & 2.6 & 0.3 & 0.6 & 0.6 & 0.8 & 12.1 \\ 
\bottomrule
\end{tabular}%

\caption{Standard deviation results in FT setup. Results are computed across 3 random seeds.}
\label{tab:std_ft}
\end{table*}

%% file: tables/app/beam_size_nft.tex
\begin{table*}[!ht]
    \centering
    \small
    \begin{threeparttable}
\begin{tabular}{@{}lcccccccccccc@{}}
\toprule
 & \multicolumn{7}{c}{\textbf{GeoQuery}} & \multicolumn{5}{c}{\textbf{SMCalFlow-CS}}\\
 \cmidrule[0.4pt](r{0.125em}){2-8}%
 \cmidrule[0.4pt](lr{0.125em}){9-13}%
 $B$ & Templ. 1 & Templ. 2 & Templ. 3 & TMCD 1 & TMCD 2 & TMCD 3 &Len. & i.i.d. & 0-C & 8-C & 16-C & 32-C \\ 
\midrule
1 & 85 & 74 & 77 & 66 & 65 & 84 & 62 & 73 & 0 & 36 & 47 & 63 \\
3 & 85 & 75 & 75 & 69 & 59 & 88 & 60 & 65 & 0 & 42 & 49 & 67 \\
5 & 84 & 76 & 72 & 69 & 64 & 87 & 60 & 64 & 1 & 44 & 51 & 68 \\
\bottomrule
\end{tabular}
\end{threeparttable}%
\caption{The effect of number of beam
candidates $B$ on accuracy in NoFT setup. Prompts are constructed using Cover-LS with $k=8$ demonstrations. Results are computed on a random subset of 100 development examples (single seed).
}
\label{tab:beam_size}
\end{table*}

%% file: tables/app/license_artifacts.tex
\begin{table*}[!ht]
\centering
\small
\begin{threeparttable}%
\begin{tabular}{@{}lll@{}}
\toprule
Artifact & License & Reference \\
\midrule
$\textbf{Models}$\\
T5 & Apache 2.0 & \href{https://huggingface.co/t5-large}{HF model card} \\
Codex & API usage policy & \href{https://beta.openai.com/docs/usage-policies}{API documentation} \\

\midrule
$\textbf{Dataset}$\\
GeoQuery & GPL 2.0 & \href{https://www.cs.utexas.edu/users/ml/nldata/geoquery.html}{Official website} \\
GeoQuery compositional splits & Apache 2.0 & \href{https://github.com/google-research/language/tree/master/language/compgen/csl/tasks/geoquery/splits}{Github repository}  \\
SMCalFlow-CS & MIT & \href{https://github.com/microsoft/compositional-generalization-span-level-attention#download-dataset}{Github repository} \\
SMCalFlow Simple & MIT & \href{https://github.com/telepathylabsai/OpenDF}{Github repository} \\
COVR-10 & MIT & \href{https://github.com/benbogin/unobserved-local-structures}{Github repository} \\

\midrule
$\textbf{Tools}$\\
AllenNLP & Apache 2.0 & \href{https://github.com/allenai/allennlp}{Github repository} \\
Rank-BM25 & Apache 2.0 & \href{https://github.com/dorianbrown/rank_bm25}{Github repository} \\
SBERT & Apache 2.0 & \href{https://github.com/UKPLab/sentence-transformers}{Github repository} \\
DPP optimization & MIT & \href{https://github.com/decile-team/submodlib}{Github repository} \\
\bottomrule
\end{tabular}%
\end{threeparttable}%
\caption{License information for all artifacts}
\label{tab:license}

\end{table*}

%% file: main.bbl
\begin{thebibliography}{47}
\expandafter\ifx\csname natexlab\endcsname\relax\def\natexlab#1{#1}\fi

\bibitem[{Andreas et~al.(2020)Andreas, Bufe, Burkett, Chen, Clausman, Crawford,
  Crim, DeLoach, Dorner, Eisner, Fang, Guo, Hall, Hayes, Hill, Ho, Iwaszuk,
  Jha, Klein, Krishnamurthy, Lanman, Liang, Lin, Lintsbakh, McGovern,
  Nisnevich, Pauls, Petters, Read, Roth, Roy, Rusak, Short, Slomin, Snyder,
  Striplin, Su, Tellman, Thomson, Vorobev, Witoszko, Wolfe, Wray, Zhang, and
  Zotov}]{andreas-etal-2020-task}
Jacob Andreas, John Bufe, David Burkett, Charles Chen, Josh Clausman, Jean
  Crawford, Kate Crim, Jordan DeLoach, Leah Dorner, Jason Eisner, Hao Fang,
  Alan Guo, David Hall, Kristin Hayes, Kellie Hill, Diana Ho, Wendy Iwaszuk,
  Smriti Jha, Dan Klein, Jayant Krishnamurthy, Theo Lanman, Percy Liang,
  Christopher~H. Lin, Ilya Lintsbakh, Andy McGovern, Aleksandr Nisnevich, Adam
  Pauls, Dmitrij Petters, Brent Read, Dan Roth, Subhro Roy, Jesse Rusak, Beth
  Short, Div Slomin, Ben Snyder, Stephon Striplin, Yu~Su, Zachary Tellman, Sam
  Thomson, Andrei Vorobev, Izabela Witoszko, Jason Wolfe, Abby Wray, Yuchen
  Zhang, and Alexander Zotov. 2020.
\newblock \href {https://doi.org/10.1162/tacl_a_00333} {Task-oriented dialogue
  as dataflow synthesis}.
\newblock \emph{Transactions of the Association for Computational Linguistics},
  8:556--571.

\bibitem[{Bogin et~al.(2022)Bogin, Gupta, and
  Berant}]{bogin-etal-2022-unobserved}
Ben Bogin, Shivanshu Gupta, and Jonathan Berant. 2022.
\newblock \href {https://aclanthology.org/2022.emnlp-main.175} {Unobserved
  local structures make compositional generalization hard}.
\newblock In \emph{Proceedings of the 2022 Conference on Empirical Methods in
  Natural Language Processing}, pages 2731--2747, Abu Dhabi, United Arab
  Emirates. Association for Computational Linguistics.

\bibitem[{Brown(2020)}]{rank_bm25}
Dorian Brown. 2020.
\newblock \href {https://doi.org/10.5281/zenodo.4520057} {{Rank-BM25: A
  Collection of BM25 Algorithms in Python}}.

\bibitem[{Brown et~al.(2020)Brown, Mann, Ryder, Subbiah, Kaplan, Dhariwal,
  Neelakantan, Shyam, Sastry, Askell, Agarwal, Herbert{-}Voss, Krueger,
  Henighan, Child, Ramesh, Ziegler, Wu, Winter, Hesse, Chen, Sigler, Litwin,
  Gray, Chess, Clark, Berner, McCandlish, Radford, Sutskever, and
  Amodei}]{Brown2020LanguageMA}
Tom~B. Brown, Benjamin Mann, Nick Ryder, Melanie Subbiah, Jared Kaplan,
  Prafulla Dhariwal, Arvind Neelakantan, Pranav Shyam, Girish Sastry, Amanda
  Askell, Sandhini Agarwal, Ariel Herbert{-}Voss, Gretchen Krueger, Tom
  Henighan, Rewon Child, Aditya Ramesh, Daniel~M. Ziegler, Jeffrey Wu, Clemens
  Winter, Christopher Hesse, Mark Chen, Eric Sigler, Mateusz Litwin, Scott
  Gray, Benjamin Chess, Jack Clark, Christopher Berner, Sam McCandlish, Alec
  Radford, Ilya Sutskever, and Dario Amodei. 2020.
\newblock \href
  {https://proceedings.neurips.cc/paper/2020/hash/1457c0d6bfcb4967418bfb8ac142f64a-Abstract.html}
  {Language models are few-shot learners}.
\newblock In \emph{Advances in Neural Information Processing Systems 33: Annual
  Conference on Neural Information Processing Systems 2020, NeurIPS 2020,
  December 6-12, 2020, virtual}.

\bibitem[{Chen et~al.(2021)Chen, Tworek, Jun, Yuan, Ponde, Kaplan, Edwards,
  Burda, Joseph, Brockman, Ray, Puri, Krueger, Petrov, Khlaaf, Sastry, Mishkin,
  Chan, Gray, Ryder, Pavlov, Power, Kaiser, Bavarian, Winter, Tillet, Such,
  Cummings, Plappert, Chantzis, Barnes, Herbert-Voss, Guss, Nichol, Babuschkin,
  Balaji, Jain, Carr, Leike, Achiam, Misra, Morikawa, Radford, Knight,
  Brundage, Murati, Mayer, Welinder, McGrew, Amodei, McCandlish, Sutskever, and
  Zaremba}]{Chen2021EvaluatingLL}
Mark Chen, Jerry Tworek, Heewoo Jun, Qiming Yuan, Henrique Ponde, Jared Kaplan,
  Harrison Edwards, Yura Burda, Nicholas Joseph, Greg Brockman, Alex Ray, Raul
  Puri, Gretchen Krueger, Michael Petrov, Heidy Khlaaf, Girish Sastry, Pamela
  Mishkin, Brooke Chan, Scott Gray, Nick Ryder, Mikhail Pavlov, Alethea Power,
  Lukasz Kaiser, Mohammad Bavarian, Clemens Winter, Philippe Tillet,
  Felipe~Petroski Such, David~W. Cummings, Matthias Plappert, Fotios Chantzis,
  Elizabeth Barnes, Ariel Herbert-Voss, William~H. Guss, Alex Nichol, Igor
  Babuschkin, S.~Arun Balaji, Shantanu Jain, Andrew Carr, Jan Leike, Joshua
  Achiam, Vedant Misra, Evan Morikawa, Alec Radford, Matthew~M. Knight, Miles
  Brundage, Mira Murati, Katie Mayer, Peter Welinder, Bob McGrew, Dario Amodei,
  Sam McCandlish, Ilya Sutskever, and Wojciech Zaremba. 2021.
\newblock \href {https://arxiv.org/abs/2107.03374} {Evaluating large language
  models trained on code}.
\newblock \emph{ArXiv preprint}, abs/2107.03374.

\bibitem[{Chen et~al.(2022)Chen, Zhong, Zha, Karypis, and
  He}]{chen-etal-2022-meta}
Yanda Chen, Ruiqi Zhong, Sheng Zha, George Karypis, and He~He. 2022.
\newblock \href {https://doi.org/10.18653/v1/2022.acl-long.53} {Meta-learning
  via language model in-context tuning}.
\newblock In \emph{Proceedings of the 60th Annual Meeting of the Association
  for Computational Linguistics (Volume 1: Long Papers)}, pages 719--730,
  Dublin, Ireland. Association for Computational Linguistics.

\bibitem[{Conklin et~al.(2021)Conklin, Wang, Smith, and
  Titov}]{conklin-etal-2021-meta}
Henry Conklin, Bailin Wang, Kenny Smith, and Ivan Titov. 2021.
\newblock \href {https://doi.org/10.18653/v1/2021.acl-long.258} {Meta-learning
  to compositionally generalize}.
\newblock In \emph{Proceedings of the 59th Annual Meeting of the Association
  for Computational Linguistics and the 11th International Joint Conference on
  Natural Language Processing (Volume 1: Long Papers)}, pages 3322--3335,
  Online. Association for Computational Linguistics.

\bibitem[{Drozdov et~al.(2022)Drozdov, Scharli, Akyuurek, Scales, Song, Chen,
  Bousquet, and Zhou}]{Drozdov2022CompositionalSP}
Andrew Drozdov, Nathanael Scharli, Ekin Akyuurek, Nathan Scales, Xinying Song,
  Xinyun Chen, Olivier Bousquet, and Denny Zhou. 2022.
\newblock \href {https://arxiv.org/abs/2209.15003} {Compositional semantic
  parsing with large language models}.
\newblock \emph{ArXiv preprint}, abs/2209.15003.

\bibitem[{Finegan-Dollak et~al.(2018)Finegan-Dollak, Kummerfeld, Zhang,
  Ramanathan, Sadasivam, Zhang, and Radev}]{finegan-dollak-etal-2018-improving}
Catherine Finegan-Dollak, Jonathan~K. Kummerfeld, Li~Zhang, Karthik Ramanathan,
  Sesh Sadasivam, Rui Zhang, and Dragomir Radev. 2018.
\newblock \href {https://doi.org/10.18653/v1/P18-1033} {Improving text-to-{SQL}
  evaluation methodology}.
\newblock In \emph{Proceedings of the 56th Annual Meeting of the Association
  for Computational Linguistics (Volume 1: Long Papers)}, pages 351--360,
  Melbourne, Australia. Association for Computational Linguistics.

\bibitem[{Finn et~al.(2017)Finn, Abbeel, and Levine}]{Finn2017ModelAgnosticMF}
Chelsea Finn, Pieter Abbeel, and Sergey Levine. 2017.
\newblock \href {http://proceedings.mlr.press/v70/finn17a.html} {Model-agnostic
  meta-learning for fast adaptation of deep networks}.
\newblock In \emph{Proceedings of the 34th International Conference on Machine
  Learning, {ICML} 2017, Sydney, NSW, Australia, 6-11 August 2017}, volume~70
  of \emph{Proceedings of Machine Learning Research}, pages 1126--1135. {PMLR}.

\bibitem[{Furrer et~al.(2020)Furrer, van Zee, Scales, and
  Scharli}]{Furrer2020CompositionalGI}
Daniel Furrer, Marc van Zee, Nathan Scales, and Nathanael Scharli. 2020.
\newblock \href {https://arxiv.org/abs/2007.08970} {Compositional
  generalization in semantic parsing: Pre-training vs. specialized
  architectures}.
\newblock \emph{ArXiv preprint}, abs/2007.08970.

\bibitem[{Gardner et~al.(2018)Gardner, Grus, Neumann, Tafjord, Dasigi, Liu,
  Peters, Schmitz, and Zettlemoyer}]{gardner-etal-2018-allennlp}
Matt Gardner, Joel Grus, Mark Neumann, Oyvind Tafjord, Pradeep Dasigi,
  Nelson~F. Liu, Matthew Peters, Michael Schmitz, and Luke Zettlemoyer. 2018.
\newblock \href {https://doi.org/10.18653/v1/W18-2501} {{A}llen{NLP}: A deep
  semantic natural language processing platform}.
\newblock In \emph{Proceedings of Workshop for {NLP} Open Source Software
  ({NLP}-{OSS})}, pages 1--6, Melbourne, Australia. Association for
  Computational Linguistics.

\bibitem[{Gupta et~al.(2022)Gupta, Singh, and
  Gardner}]{gupta-etal-2022-structurally}
Shivanshu Gupta, Sameer Singh, and Matt Gardner. 2022.
\newblock \href {https://aclanthology.org/2022.findings-emnlp.365}
  {Structurally diverse sampling for sample-efficient training and
  comprehensive evaluation}.
\newblock In \emph{Findings of the Association for Computational Linguistics:
  EMNLP 2022}, pages 4966--4979, Abu Dhabi, United Arab Emirates. Association
  for Computational Linguistics.

\bibitem[{Herzig and Berant(2019)}]{herzig-berant-2019-dont}
Jonathan Herzig and Jonathan Berant. 2019.
\newblock \href {https://doi.org/10.18653/v1/D19-1394} {Don{'}t paraphrase,
  detect! rapid and effective data collection for semantic parsing}.
\newblock In \emph{Proceedings of the 2019 Conference on Empirical Methods in
  Natural Language Processing and the 9th International Joint Conference on
  Natural Language Processing (EMNLP-IJCNLP)}, pages 3810--3820, Hong Kong,
  China. Association for Computational Linguistics.

\bibitem[{Hosseini et~al.(2022)Hosseini, Vani, Bahdanau, Sordoni, and
  Courville}]{hosseini-etal-2022-compositional}
Arian Hosseini, Ankit Vani, Dzmitry Bahdanau, Alessandro Sordoni, and Aaron
  Courville. 2022.
\newblock \href {https://aclanthology.org/2022.blackboxnlp-1.22} {On the
  compositional generalization gap of in-context learning}.
\newblock In \emph{Proceedings of the Fifth BlackboxNLP Workshop on Analyzing
  and Interpreting Neural Networks for NLP}, pages 272--280, Abu Dhabi, United
  Arab Emirates (Hybrid). Association for Computational Linguistics.

\bibitem[{Hupkes et~al.(2022)Hupkes, Giulianelli, Dankers, Artetxe, Elazar,
  Pimentel, Christodoulopoulos, Lasri, Saphra, Sinclair, Ulmer, Schottmann,
  Batsuren, Sun, Sinha, Khalatbari, Ryskina, Frieske, Cotterell, and
  Jin}]{hupkes2022taxonomy}
Dieuwke Hupkes, Mario Giulianelli, Verna Dankers, Mikel Artetxe, Yanai Elazar,
  Tiago Pimentel, Christos Christodoulopoulos, Karim Lasri, Naomi Saphra,
  Arabella Sinclair, Dennis Ulmer, Florian Schottmann, Khuyagbaatar Batsuren,
  Kaiser Sun, Koustuv Sinha, Leila Khalatbari, Maria Ryskina, Rita Frieske,
  Ryan Cotterell, and Zhijing Jin. 2022.
\newblock \href {https://arxiv.org/abs/2210.03050} {State-of-the-art
  generalisation research in {NLP}: a taxonomy and review}.
\newblock \emph{ArXiv preprint}, abs/2210.03050.

\bibitem[{Kaushal et~al.(2022)Kaushal, Ramakrishnan, and
  Iyer}]{Kaushal2022SubmodlibAS}
Vishal Kaushal, Ganesh Ramakrishnan, and Rishabh~K. Iyer. 2022.
\newblock \href {https://arxiv.org/abs/2202.10680} {Submodlib: A submodular
  optimization library}.
\newblock \emph{ArXiv preprint}, abs/2202.10680.

\bibitem[{Keysers et~al.(2020)Keysers, Sch{\"{a}}rli, Scales, Buisman, Furrer,
  Kashubin, Momchev, Sinopalnikov, Stafiniak, Tihon, Tsarkov, Wang, van Zee,
  and Bousquet}]{keysers2020measuring}
Daniel Keysers, Nathanael Sch{\"{a}}rli, Nathan Scales, Hylke Buisman, Daniel
  Furrer, Sergii Kashubin, Nikola Momchev, Danila Sinopalnikov, Lukasz
  Stafiniak, Tibor Tihon, Dmitry Tsarkov, Xiao Wang, Marc van Zee, and Olivier
  Bousquet. 2020.
\newblock \href {https://openreview.net/forum?id=SygcCnNKwr} {Measuring
  compositional generalization: {A} comprehensive method on realistic data}.
\newblock In \emph{8th International Conference on Learning Representations,
  {ICLR} 2020, Addis Ababa, Ethiopia, April 26-30, 2020}. OpenReview.net.

\bibitem[{Kulesza and Taskar(2012)}]{Kulesza2012DeterminantalPP}
Alex Kulesza and Ben Taskar. 2012.
\newblock \href {https://doi.org/10.1561/2200000044} {Determinantal point
  processes for machine learning}.
\newblock \emph{Foundations and Trends® in Machine Learning},
  5(2–3):123--286.

\bibitem[{Lake(2019)}]{Lake2019CompositionalGT}
Brenden~M. Lake. 2019.
\newblock \href
  {https://proceedings.neurips.cc/paper/2019/hash/f4d0e2e7fc057a58f7ca4a391f01940a-Abstract.html}
  {Compositional generalization through meta sequence-to-sequence learning}.
\newblock In \emph{Advances in Neural Information Processing Systems 32: Annual
  Conference on Neural Information Processing Systems 2019, NeurIPS 2019,
  December 8-14, 2019, Vancouver, BC, Canada}, pages 9788--9798.

\bibitem[{Lake and Baroni(2018)}]{pmlr-v80-lake18a}
Brenden~M. Lake and Marco Baroni. 2018.
\newblock \href {http://proceedings.mlr.press/v80/lake18a.html} {Generalization
  without systematicity: On the compositional skills of sequence-to-sequence
  recurrent networks}.
\newblock In \emph{Proceedings of the 35th International Conference on Machine
  Learning, {ICML} 2018, Stockholmsm{\"{a}}ssan, Stockholm, Sweden, July 10-15,
  2018}, volume~80 of \emph{Proceedings of Machine Learning Research}, pages
  2879--2888. {PMLR}.

\bibitem[{Liu et~al.(2022)Liu, Shen, Zhang, Dolan, Carin, and
  Chen}]{liu-etal-2022-makes}
Jiachang Liu, Dinghan Shen, Yizhe Zhang, Bill Dolan, Lawrence Carin, and Weizhu
  Chen. 2022.
\newblock \href {https://doi.org/10.18653/v1/2022.deelio-1.10} {What makes good
  in-context examples for {GPT}-3?}
\newblock In \emph{Proceedings of Deep Learning Inside Out (DeeLIO 2022): The
  3rd Workshop on Knowledge Extraction and Integration for Deep Learning
  Architectures}, pages 100--114, Dublin, Ireland and Online. Association for
  Computational Linguistics.

\bibitem[{Loshchilov and Hutter(2019)}]{Loshchilov2017DecoupledWD}
Ilya Loshchilov and Frank Hutter. 2019.
\newblock \href {https://openreview.net/forum?id=Bkg6RiCqY7} {Decoupled weight
  decay regularization}.
\newblock In \emph{7th International Conference on Learning Representations,
  {ICLR} 2019, New Orleans, LA, USA, May 6-9, 2019}. OpenReview.net.

\bibitem[{Meister et~al.(2021)Meister, Forster, and
  Cotterell}]{meister-etal-2021-determinantal}
Clara Meister, Martina Forster, and Ryan Cotterell. 2021.
\newblock \href {https://doi.org/10.18653/v1/2021.acl-long.512} {Determinantal
  beam search}.
\newblock In \emph{Proceedings of the 59th Annual Meeting of the Association
  for Computational Linguistics and the 11th International Joint Conference on
  Natural Language Processing (Volume 1: Long Papers)}, pages 6551--6562,
  Online. Association for Computational Linguistics.

\bibitem[{Meron(2022)}]{meron-2022-simplifying}
Joram Meron. 2022.
\newblock \href {https://aclanthology.org/2022.isa-1.11} {Simplifying semantic
  annotations of {SMC}al{F}low}.
\newblock In \emph{Proceedings of the 18th Joint ACL - ISO Workshop on
  Interoperable Semantic Annotation within LREC2022}, pages 81--85, Marseille,
  France. European Language Resources Association.

\bibitem[{Min et~al.(2021)Min, Lee, Chang, Toutanova, and
  Hajishirzi}]{min-etal-2021-joint}
Sewon Min, Kenton Lee, Ming-Wei Chang, Kristina Toutanova, and Hannaneh
  Hajishirzi. 2021.
\newblock \href {https://doi.org/10.18653/v1/2021.emnlp-main.560} {Joint
  passage ranking for diverse multi-answer retrieval}.
\newblock In \emph{Proceedings of the 2021 Conference on Empirical Methods in
  Natural Language Processing}, pages 6997--7008, Online and Punta Cana,
  Dominican Republic. Association for Computational Linguistics.

\bibitem[{Min et~al.(2022)Min, Lewis, Zettlemoyer, and
  Hajishirzi}]{min-etal-2022-metaicl}
Sewon Min, Mike Lewis, Luke Zettlemoyer, and Hannaneh Hajishirzi. 2022.
\newblock \href {https://doi.org/10.18653/v1/2022.naacl-main.201} {{M}eta{ICL}:
  Learning to learn in context}.
\newblock In \emph{Proceedings of the 2022 Conference of the North American
  Chapter of the Association for Computational Linguistics: Human Language
  Technologies}, pages 2791--2809, Seattle, United States. Association for
  Computational Linguistics.

\bibitem[{Nandigam et~al.(2022)Nandigam, Rayaprolu, and
  Shrivastava}]{nandigam-etal-2022-diverse}
Poojitha Nandigam, Nikhil Rayaprolu, and Manish Shrivastava. 2022.
\newblock \href {https://aclanthology.org/2022.coling-1.194} {Diverse
  multi-answer retrieval with determinantal point processes}.
\newblock In \emph{Proceedings of the 29th International Conference on
  Computational Linguistics}, pages 2220--2225, Gyeongju, Republic of Korea.
  International Committee on Computational Linguistics.

\bibitem[{Narayan et~al.(2022)Narayan, Sim{\~o}es, Zhao, Maynez, Das, Collins,
  and Lapata}]{narayan-etal-2022-well}
Shashi Narayan, Gon{\c{c}}alo Sim{\~o}es, Yao Zhao, Joshua Maynez, Dipanjan
  Das, Michael Collins, and Mirella Lapata. 2022.
\newblock \href {https://doi.org/10.18653/v1/2022.acl-long.94} {A well-composed
  text is half done! composition sampling for diverse conditional generation}.
\newblock In \emph{Proceedings of the 60th Annual Meeting of the Association
  for Computational Linguistics (Volume 1: Long Papers)}, pages 1319--1339,
  Dublin, Ireland. Association for Computational Linguistics.

\bibitem[{Oren et~al.(2021)Oren, Herzig, and Berant}]{oren-etal-2021-finding}
Inbar Oren, Jonathan Herzig, and Jonathan Berant. 2021.
\newblock \href {https://doi.org/10.18653/v1/2021.emnlp-main.843} {Finding
  needles in a haystack: Sampling structurally-diverse training sets from
  synthetic data for compositional generalization}.
\newblock In \emph{Proceedings of the 2021 Conference on Empirical Methods in
  Natural Language Processing}, pages 10793--10809, Online and Punta Cana,
  Dominican Republic. Association for Computational Linguistics.

\bibitem[{Ouyang et~al.(2022)Ouyang, Wu, Jiang, Almeida, Wainwright, Mishkin,
  Zhang, Agarwal, Slama, Ray, Schulman, Hilton, Kelton, Miller, Simens, Askell,
  Welinder, Christiano, Leike, and Lowe}]{Ouyang2022TrainingLM}
Long Ouyang, Jeff Wu, Xu~Jiang, Diogo Almeida, Carroll~L. Wainwright, Pamela
  Mishkin, Chong Zhang, Sandhini Agarwal, Katarina Slama, Alex Ray, John
  Schulman, Jacob Hilton, Fraser Kelton, Luke~E. Miller, Maddie Simens, Amanda
  Askell, Peter Welinder, Paul~Francis Christiano, Jan Leike, and Ryan~J. Lowe.
  2022.
\newblock \href {https://arxiv.org/abs/2203.02155} {Training language models to
  follow instructions with human feedback}.
\newblock \emph{ArXiv preprint}, abs/2203.02155.

\bibitem[{Pasupat et~al.(2021)Pasupat, Zhang, and
  Guu}]{pasupat-etal-2021-controllable}
Panupong Pasupat, Yuan Zhang, and Kelvin Guu. 2021.
\newblock \href {https://doi.org/10.18653/v1/2021.emnlp-main.607} {Controllable
  semantic parsing via retrieval augmentation}.
\newblock In \emph{Proceedings of the 2021 Conference on Empirical Methods in
  Natural Language Processing}, pages 7683--7698, Online and Punta Cana,
  Dominican Republic. Association for Computational Linguistics.

\bibitem[{Pedregosa et~al.(2011)Pedregosa, Varoquaux, Gramfort, Michel,
  Thirion, Grisel, Blondel, Prettenhofer, Weiss, Dubourg, Vanderplas, Passos,
  Cournapeau, Brucher, Perrot, and Duchesnay}]{scikit-learn}
F.~Pedregosa, G.~Varoquaux, A.~Gramfort, V.~Michel, B.~Thirion, O.~Grisel,
  M.~Blondel, P.~Prettenhofer, R.~Weiss, V.~Dubourg, J.~Vanderplas, A.~Passos,
  D.~Cournapeau, M.~Brucher, M.~Perrot, and E.~Duchesnay. 2011.
\newblock Scikit-learn: Machine learning in {P}ython.
\newblock \emph{Journal of Machine Learning Research}, 12:2825--2830.

\bibitem[{Qiu et~al.(2022{\natexlab{a}})Qiu, Shaw, Pasupat, Nowak, Linzen, Sha,
  and Toutanova}]{qiu-etal-2022-improving}
Linlu Qiu, Peter Shaw, Panupong Pasupat, Pawel Nowak, Tal Linzen, Fei Sha, and
  Kristina Toutanova. 2022{\natexlab{a}}.
\newblock \href {https://doi.org/10.18653/v1/2022.naacl-main.323} {Improving
  compositional generalization with latent structure and data augmentation}.
\newblock In \emph{Proceedings of the 2022 Conference of the North American
  Chapter of the Association for Computational Linguistics: Human Language
  Technologies}, pages 4341--4362, Seattle, United States. Association for
  Computational Linguistics.

\bibitem[{Qiu et~al.(2022{\natexlab{b}})Qiu, Shaw, Pasupat, Shi, Herzig,
  Pitler, Sha, and Toutanova}]{qiu-etal-2022-evaluating}
Linlu Qiu, Peter Shaw, Panupong Pasupat, Tianze Shi, Jonathan Herzig, Emily
  Pitler, Fei Sha, and Kristina Toutanova. 2022{\natexlab{b}}.
\newblock \href {https://aclanthology.org/2022.emnlp-main.624} {Evaluating the
  impact of model scale for compositional generalization in semantic parsing}.
\newblock In \emph{Proceedings of the 2022 Conference on Empirical Methods in
  Natural Language Processing}, pages 9157--9179, Abu Dhabi, United Arab
  Emirates. Association for Computational Linguistics.

\bibitem[{Raffel et~al.(2020)Raffel, Shazeer, Roberts, Lee, Narang, Matena,
  Zhou, Li, and Liu}]{Raffel2020ExploringTL}
Colin Raffel, Noam Shazeer, Adam Roberts, Katherine Lee, Sharan Narang, Michael
  Matena, Yanqi Zhou, Wei Li, and Peter~J. Liu. 2020.
\newblock \href {http://jmlr.org/papers/v21/20-074.html} {Exploring the limits
  of transfer learning with a unified text-to-text transformer}.
\newblock \emph{J. Mach. Learn. Res.}, 21:140:1--140:67.

\bibitem[{Reimers and Gurevych(2019)}]{reimers-gurevych-2019-sentence}
Nils Reimers and Iryna Gurevych. 2019.
\newblock \href {https://doi.org/10.18653/v1/D19-1410} {Sentence-{BERT}:
  Sentence embeddings using {S}iamese {BERT}-networks}.
\newblock In \emph{Proceedings of the 2019 Conference on Empirical Methods in
  Natural Language Processing and the 9th International Joint Conference on
  Natural Language Processing (EMNLP-IJCNLP)}, pages 3982--3992, Hong Kong,
  China. Association for Computational Linguistics.

\bibitem[{Robertson and Zaragoza(2009)}]{Robertson2009ThePR}
Stephen Robertson and Hugo Zaragoza. 2009.
\newblock \href {https://doi.org/10.1561/1500000019} {The probabilistic
  relevance framework: Bm25 and beyond}.
\newblock \emph{Foundations and Trends® in Information Retrieval},
  3(4):333--389.

\bibitem[{Rubin et~al.(2022)Rubin, Herzig, and
  Berant}]{rubin-etal-2022-learning}
Ohad Rubin, Jonathan Herzig, and Jonathan Berant. 2022.
\newblock \href {https://doi.org/10.18653/v1/2022.naacl-main.191} {Learning to
  retrieve prompts for in-context learning}.
\newblock In \emph{Proceedings of the 2022 Conference of the North American
  Chapter of the Association for Computational Linguistics: Human Language
  Technologies}, pages 2655--2671, Seattle, United States. Association for
  Computational Linguistics.

\bibitem[{Shaw et~al.(2021)Shaw, Chang, Pasupat, and
  Toutanova}]{shaw-etal-2021-compositional}
Peter Shaw, Ming-Wei Chang, Panupong Pasupat, and Kristina Toutanova. 2021.
\newblock \href {https://doi.org/10.18653/v1/2021.acl-long.75} {Compositional
  generalization and natural language variation: Can a semantic parsing
  approach handle both?}
\newblock In \emph{Proceedings of the 59th Annual Meeting of the Association
  for Computational Linguistics and the 11th International Joint Conference on
  Natural Language Processing (Volume 1: Long Papers)}, pages 922--938, Online.
  Association for Computational Linguistics.

\bibitem[{Su et~al.(2022)Su, Kasai, Wu, Shi, Wang, Xin, Zhang, Ostendorf,
  Zettlemoyer, Smith, and Yu}]{Su2022SelectiveAM}
Hongjin Su, Jungo Kasai, Chen~Henry Wu, Weijia Shi, Tianlu Wang, Jiayi Xin, Rui
  Zhang, Mari Ostendorf, Luke Zettlemoyer, Noah~A. Smith, and Tao Yu. 2022.
\newblock \href {https://arxiv.org/abs/2209.01975} {Selective annotation makes
  language models better few-shot learners}.
\newblock \emph{ArXiv preprint}, abs/2209.01975.

\bibitem[{Tang and Mooney(2001)}]{Tang2001UsingMC}
Lappoon~R. Tang and Raymond~J. Mooney. 2001.
\newblock \href {https://link.springer.com/chapter/10.1007/3-540-44795-4_40}
  {Using multiple clause constructors in inductive logic programming for
  semantic parsing}.
\newblock In \emph{ECML}.

\bibitem[{Wang et~al.(2022)Wang, Xu, Fang, Liu, Sun, Xu, Zhu, and
  Zeng}]{wang-etal-2022-training}
Shuohang Wang, Yichong Xu, Yuwei Fang, Yang Liu, Siqi Sun, Ruochen Xu,
  Chenguang Zhu, and Michael Zeng. 2022.
\newblock \href {https://doi.org/10.18653/v1/2022.acl-long.226} {Training data
  is more valuable than you think: A simple and effective method by retrieving
  from training data}.
\newblock In \emph{Proceedings of the 60th Annual Meeting of the Association
  for Computational Linguistics (Volume 1: Long Papers)}, pages 3170--3179,
  Dublin, Ireland. Association for Computational Linguistics.

\bibitem[{Ye et~al.(2022)Ye, Iyer, Celikyilmaz, Stoyanov, Durrett, and
  Pasunuru}]{Ye2022ComplementaryEF}
Xi~Ye, Srini Iyer, Asli Celikyilmaz, Ves Stoyanov, Greg Durrett, and Ramakanth
  Pasunuru. 2022.
\newblock \href {https://arxiv.org/abs/2211.13892} {Complementary explanations
  for effective in-context learning}.
\newblock \emph{ArXiv preprint}, abs/2211.13892.

\bibitem[{Yin et~al.(2021)Yin, Fang, Neubig, Pauls, Platanios, Su, Thomson, and
  Andreas}]{yin-etal-2021-compositional}
Pengcheng Yin, Hao Fang, Graham Neubig, Adam Pauls, Emmanouil~Antonios
  Platanios, Yu~Su, Sam Thomson, and Jacob Andreas. 2021.
\newblock \href {https://doi.org/10.18653/v1/2021.naacl-main.225}
  {Compositional generalization for neural semantic parsing via span-level
  supervised attention}.
\newblock In \emph{Proceedings of the 2021 Conference of the North American
  Chapter of the Association for Computational Linguistics: Human Language
  Technologies}, pages 2810--2823, Online. Association for Computational
  Linguistics.

\bibitem[{Zelle and Mooney(1996)}]{Zelle1996LearningTP}
John~M. Zelle and Raymond~J. Mooney. 1996.
\newblock \href {https://dl.acm.org/doi/10.5555/1864519.1864543} {Learning to
  parse database queries using inductive logic programming}.
\newblock In \emph{AAAI/IAAI, Vol. 2}.

\bibitem[{Zemlyanskiy et~al.(2022)Zemlyanskiy, de~Jong, Ainslie, Pasupat, Shaw,
  Qiu, Sanghai, and Sha}]{zemlyanskiy-etal-2022-generate}
Yury Zemlyanskiy, Michiel de~Jong, Joshua Ainslie, Panupong Pasupat, Peter
  Shaw, Linlu Qiu, Sumit Sanghai, and Fei Sha. 2022.
\newblock \href {https://aclanthology.org/2022.coling-1.438}
  {Generate-and-retrieve: Use your predictions to improve retrieval for
  semantic parsing}.
\newblock In \emph{Proceedings of the 29th International Conference on
  Computational Linguistics}, pages 4946--4951, Gyeongju, Republic of Korea.
  International Committee on Computational Linguistics.

\end{thebibliography}
